\documentclass[10pt,journal,compsoc]{IEEEtran}
\usepackage{graphicx} %
    
\usepackage{amsmath}
\usepackage{amssymb}
\usepackage{booktabs}
\usepackage{multirow}
\usepackage{xcolor}
\usepackage{bbm}
\usepackage{array}
\usepackage{ragged2e} 
\usepackage{arydshln}
\usepackage{scalerel,stackengine}

\definecolor{citecolor}{HTML}{0071bc}
\usepackage[pagebackref=false,breaklinks=true,colorlinks,citecolor=citecolor,bookmarks=false]{hyperref}

\newlength\savewidth\newcommand\shline{\noalign{\global\savewidth\arrayrulewidth
  \global\arrayrulewidth 1pt}\hline\noalign{\global\arrayrulewidth\savewidth}}

\makeatletter\renewcommand\paragraph{\@startsection{paragraph}{4}{\z@}
  {.5em \@plus1ex \@minus.2ex}{-.5em}{\normalfont\normalsize\bfseries}}\makeatother

\definecolor{cyan}{RGB}{0,204,204}
\newcommand{\sDelta}{\Delta\!\!\!\!\Delta}
\newcommand\wh[1]{\hstretch{2}{\hat{\hstretch{.5}{#1}}}}
\def\cL{{\cal L}}

\ifCLASSOPTIONcompsoc
  \usepackage[nocompress]{cite}
\else
  \usepackage{cite}
\fi

\newcommand{\ie}{\textit{i}.\textit{e}.}
\newcommand{\eg}{\textit{e}.\textit{g}.}

\begin{document}

\title{Pruning Self-attentions into Convolutional Layers in Single Path}

\author{Haoyu~He, Jianfei~Cai, Jing~Liu, Zizheng~Pan, Jing~Zhang, Dacheng~Tao, and Bohan~Zhuang
\IEEEcompsocitemizethanks{\IEEEcompsocthanksitem H.~He, J.~Cai, J.~Liu, Z.~Pan, and B.~Zhuang are with the Department of Data Science and AI, Faculty of IT, Monash University, Australia.
E-mail: \{Haoyu.He, Jianfei.Cai, Jing.Liu1, Zizheng.Pan, Bohan.Zhuang\}@monash.edu.
\protect

\IEEEcompsocthanksitem J.~Zhang is with the Faculty of Engineering, The University of Sydney, Australia. E-mail: Jing.Zhang1@sydney.edu.au.
\protect
\IEEEcompsocthanksitem D. Tao is with the School of Computer Science and Engineering at Nanyang Technological University, Singapore. Email: dacheng.tao@ntu.edu.sg.
\protect
\IEEEcompsocthanksitem Bohan Zhuang is the corresponding author.

}%

}

\markboth{}%
{Shell \MakeLowercase{\textit{et al.}}: Bare Demo of IEEEtran.cls for Computer Society Journals}

\IEEEtitleabstractindextext{%
\begin{abstract}
\justifying
Vision Transformers (ViTs) have achieved impressive performance over various computer vision tasks. However, modeling global correlations with multi-head self-attention (MSA) layers leads to two widely recognized issues: the massive computational resource consumption and the lack of intrinsic inductive bias for modeling local visual patterns. To solve both issues, we devise a simple yet effective method named \textbf{S}ingle-\textbf{P}ath \textbf{Vi}sion \textbf{T}ransformer pruning (SPViT), to efficiently and automatically compress the pre-trained ViTs into compact models with proper locality added. Specifically, we first propose a novel weight-sharing scheme between MSA and convolutional operations, delivering a single-path space to encode all candidate operations.
In this way, we cast the operation search problem as finding which subset of parameters to use in each MSA layer, which significantly reduces the computational cost and optimization difficulty, and the convolution kernels can be well initialized using pre-trained MSA parameters.
Relying on the single-path space, we introduce learnable binary gates to encode the operation choices in MSA layers. Similarly, we further employ learnable gates to encode the fine-grained MLP expansion ratios of FFN layers. In this way, our SPViT optimizes the learnable gates to automatically explore from a vast and unified search space and flexibly adjust the MSA-FFN pruning proportions for each individual dense model. We conduct extensive experiments on two representative ViTs showing that our SPViT achieves a new SOTA for pruning on ImageNet-1k. For example, our SPViT can trim 52.0\% FLOPs for DeiT-B and get an impressive 0.6\% top-1 accuracy gain simultaneously. The source code is available at \url{https://github.com/ziplab/SPViT}.

\end{abstract}

}

\maketitle

\IEEEdisplaynontitleabstractindextext

\IEEEpeerreviewmaketitle

\section{Introduction}\label{sec:introduction}

\IEEEPARstart{V}{ision} Transformers~\cite{vit,hvt,swin}  have attracted substantial research interests and become one of the dominant backbones in various image recognition tasks, such as classification~\cite{touvron2021training,swin,hvt,zhang2022vsa}, segmentation~\cite{zheng2021rethinking,cheng2021per,wang2021max} and detection~\cite{carion2020end,zhu2021deformable,gao2021fast}.

\begin{figure}[t]
  \centering
\includegraphics[width=\linewidth]{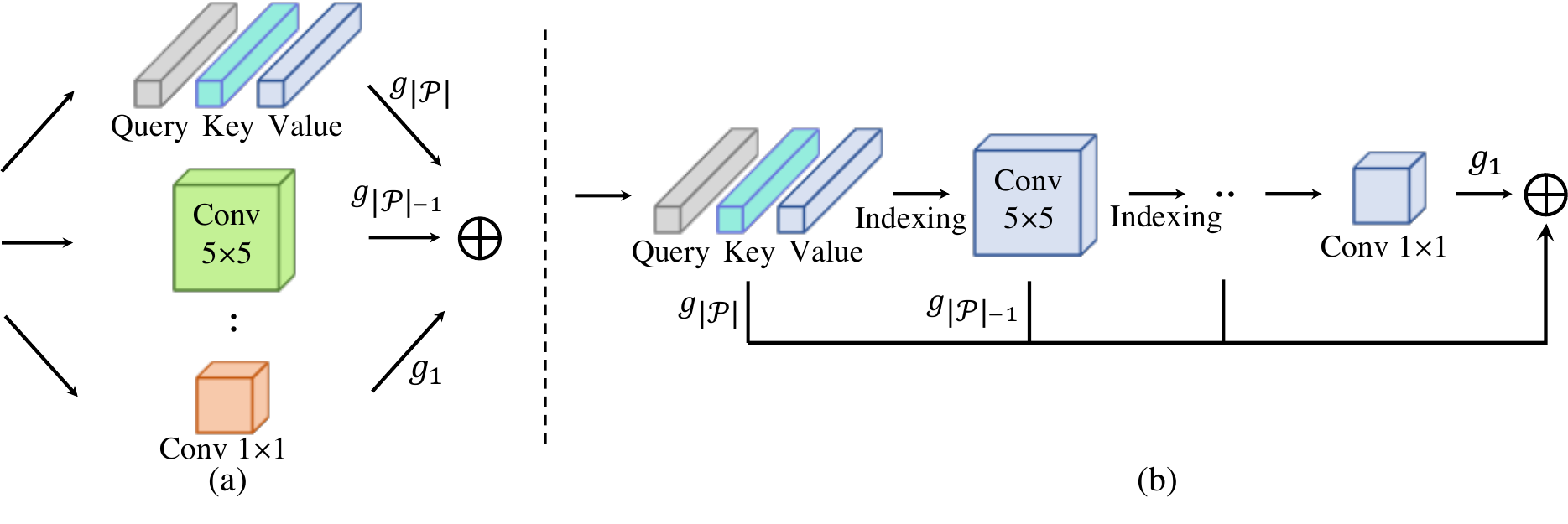}
  \caption{Multi-path vs. single-path search space. (a) Multi-path search space. MSA and different types of convolutional operations are added as separate trainable paths, leading to an expensive search cost.
  $\mathcal{P}$ and $g_p$ denote the ordered operation set and the learnable gate for the $p$-th operation, respectively. (b) Single-path search space. We propose to formulate searching as a subset selection problem where the outputs of the convolutional operations are directly indexed from the MSA intermediate results, thereby easing the optimization difficulty and reducing the search cost together with the number of trainable parameters.}
  \vspace{-0.5em}
  \label{fig:multi_vs_single}
\end{figure}

Despite the huge excitement from the recent development, two limitations of ViTs introduced by multi-head self-attention layers~\cite{transformer} have been recognized. Firstly, a well-known concern with MSA layers is the quadratic time and memory complexity, hindering the development and deployment of ViTs at scale, especially for modeling long sequences. To this end, post-training pruning methods~\cite{zhu2021visual,yang2021nvit} prune the less important ViT components from pre-trained ViT models to adapt ViTs to more resource-limited scenarios. Another fundamental problem is that MSA layers lack a locality modeling mechanism for encoding local information~\cite{li2021localvit}, which is essential for handling low-level image patterns, \eg, edges and shapes. In this case, prior arts propose to introduce inductive bias by inserting convolutional layers in ViTs with various heuristics, \eg, inside feed-forward networks (FFNs)~\cite{li2021localvit,guo2021cmt}, before the ViT encoder~\cite{xiao2021early} or before the MSA layers~\cite{xu2021vitae}. One question is consequently asked: \textbf{\emph{How to automatically prune pre-trained ViT models into efficient ones while adding proper locality at the same time}}? Recently, several neural architecture search (NAS) methods~\cite{chen2021glit,xu2021bert,li2021bossnas} propose to include candidate MSA and convolutional operations separately into the search space, thereby deriving efficient ViT models with locality. In this case, different operations are maintained as different paths and the NAS problem is solved by sampling a path from the multi-path space, as shown in Figure~\ref{fig:multi_vs_single} (a). However, different operations are trained independently, which is computationally expensive yet unnecessary.

In this paper, we propose a simple yet effective method named \textbf{S}ingle-\textbf{P}ath \textbf{Vi}sion \textbf{T}ransformer (SPViT), which automatically prunes pre-trained ViTs into compact ones with proper locality introduced. Critically, our SPViT exhibits low search costs. To be specific, inspired by the findings in~\cite{cordonnier2019relationship} that MSA layers have the capacity for modeling local regions, we first develop a weight-sharing scheme between MSA and convolutional operations, which expresses convolutional operations with a subset of the MSA parameters. With the weight-sharing scheme, we further devise a novel \textbf{\emph{single-path}} search space to encapsulate all the candidate operations (convolutions and MSA) into a single MSA per layer, as illustrated in Figure~\ref{fig:multi_vs_single} (b). Therefore, we can directly derive the outputs of the convolutional operations by indexing MSA intermediate results and obtain good initialization of the convolution kernels using pre-trained MSA parameters. Instead of choosing a path in the multi-path space \cite{chen2021glit,xu2021bert,li2021bossnas}, our SPViT formulates the search problem as finding which subset of weights to use in each MSA layer, which significantly reduces the search cost and optimization difficulty. 

We next propose a novel approach to search for compact and efficient architectures for ViTs using NAS. Our approach integrates pruning MSA and FFN layers within a unified search space. Specifically, we first introduce learnable binary gates into each MSA layer to encode the choice of operations (convolutions and MSA) based on our single-path search space. Additionally, considering that FFN layers consume a significant portion of computations (\eg, 11.1G out of 17.5G Mult-Adds for DeiT-B model~\cite{touvron2021training} when processing 224$\times$224 images), we incorporate pruning FFN layers. To jointly prune MSA and FFNs in a unified search space, we also introduce learnable binary gates to each of the FFN hidden dimensions. Our method then automatically optimizes these learnable gates and explores a wide range of efficient architectures with flexibly adjusted MSA-FFN pruning proportions for each individual dense model. After the search phase, we follow pruning literature~\cite{yu2022unified,chen2021chasing,molchanov2016pruning} to fine-tune the searched architectures and deliver compact models.

Our main contributions can be summarized as follows:

\begin{itemize}
\itemsep -0.015cm
    \item We propose a novel weight-sharing scheme between the MSA and convolutional operations, which enables encoding all candidate operations into an MSA layer in a single-path search space. We then cast the search problem as finding the subset of MSA parameters, thereby significantly reducing the search cost and optimization difficulty. 
    \item Following the
    single-path search strategy, 
    we propose SPViT that automatically prunes the costly and global
    MSA operations into the lightweight and local convolutional operations as well as searching for fine-grained MLP expansion ratios under desired efficiency constraints.
    \item
    Extensive experiments
on ImageNet-1k~\cite{russakovsky2015imagenet} show that our SPViT derives favorable pruning performance when pruning DeiT~\cite{touvron2021training}, and Swin~\cite{swin} models, achieving SOTA pruning performance. For instance, our SPViT variants with knowledge distillation achieve 23.1\%, 28.3\%, and 52.0\% FLOPs reduction for DeiT-Ti, DeiT-S, and DeiT-B models with 1.0\%, 0.4\%, 0.6\% top-1 accuracy gain over the uncompressed model, respectively. We also make several interesting observations on the searched architectures revealing the architecture preferences when pruning ViTs.
\end{itemize}
\section{Related Work}\label{subsec:related_work}
\subsection{Pruning Transformers}\label{subsec:related_work_pruning} Model pruning is a dominant approach to alleviate the high computational cost for Transformers. Prior arts for pruning Transformers can be roughly categorized into \emph{token compression} and \emph{module pruning}. 

\emph{Token compression} methods focus on pruning less-important tokens~\cite{pan2021ia,rao2021dynamicvit} or merging the similar ones~\cite{bolya2022token,wei2023joint}, \eg, DynamicViT~\cite{rao2021dynamicvit} progressively predicts binary decision masks and prunes the masked uninformative tokens. Although promising, these methods keep or even enlarge the original model size, leading to substantial storage costs. Furthermore, with a reduced sequence length, it is challenging to extend token compression to dense prediction tasks, \eg, segmentation, and detection. 
            
\emph{Module pruning} methods prune the insignificant Transformer modules, \eg, heads in the MSA layers~\cite{michel2019sixteen,behnke2020losing}, channels for the linear projections~\cite{mao2021tprune,li2020efficient,bartoldson2019generalization} and weight neurons~\cite{chen2020lottery,prasanna2020bert}. Recently, to achieve higher FLOPs-accuracy trade-offs, several works~\cite{yu2022unified, hou2021multi, chen2021chasing} enlarge the pruning space to include multiple ViT modules. For instance, MDC~\cite{hou2021multi} jointly prunes tokens and modules and UVC~\cite{yu2022unified} integrates pruning the output projections, the MSA heads, and transformer blocks, simultaneously. Another line of research~\cite{yang2021nvit, yu2023x} seeks better compact architectures with improved pruning metrics under the existing pruning spaces, such as a global hessian-based group importance~\cite{yang2021nvit} or a per-class explainable pruning metric~\cite{yu2023x}.

\emph{Our work falls into the category that prunes network modules.} In contrast to previous works~\cite{yu2022unified, hou2021multi, chen2021chasing, yang2021nvit, yu2023x} that prune MSA heads as a standard practice, our work prunes the superfluous global correlations in MSA layers and turns them into convolutional layers. In this way, instead of getting compact ViT architectures in~\cite{yu2022unified, hou2021multi, chen2021chasing, yang2021nvit, yu2023x}, we derive hybrid architectures with locality added that largely outperform head pruning (Section~\ref{subsec:locality}). Moreover, based on our novel weight-sharing scheme that expresses convolutions with a subset of MSA parameters, we form a single-path search space rather than a multi-path one~\cite{xu2021bert} to enjoy a low search cost (Section~\ref{subsec:abl_single_path}). However, to avoid overly complicating the overall framework, we only prune two components: the global correlation in MSAs and the unimportant hidden dimensions in FFNs. Given the orthogonal nature of our method to pruning more ViT components and better pruning metrics, we take integrating them into our method as a future work.

It’s worth mentioning that some methods start pruning from a randomly initialized model with sparsity constraint and perform weight removal during the training itself, \eg, S$^2$ViTE~\cite{chen2021chasing} employs a prune-and-grow strategy to explore a larger pruning space under the lottery ticket hypothesis \cite{evci2020rigging}. Although its performance is not constrained by the pre-trained models, it requires a longer training schedule and is more computationally expensive. Compared to S$^2$ViTE, our method can quickly slim the pre-trained dense ViT models into compact ones and achieve even better performance at the same time when incorporated with knowledge distillation~\cite{hinton2015distilling}.

\vspace{-1em}
\subsection{Convolutional vs. Self-Attention Layers}
Our work is motivated by recent studies that explore the properties of convolutional and self-attention layers. For instance, convolutional layers show a strong capability for extracting local texture cues with convolutional inductive bias~\cite{geirhos2018imagenet,brendel2019approximating}. On the other hand, self-attention layers tend to emphasize object shapes by modeling global correlations~\cite{naseer2021intriguing,caron2021emerging}. 

To enjoy merits from both operations, one line of work~\cite{li2021localvit,xu2021vitae,mehta2021mobilevit} treats convolutional and self-attention layers independently and combines them in different ways. For example, for better performance, various research
efforts have been made to insert convolutional layers in ViT blocks~\cite{li2021localvit,xu2021vitae,zhang2023lite,zhang2023completionformer}, put convolutional layers at different stages~\cite{gao2022convmae,touvron2021augmenting}, or stack self-attentions on top of CNNs~\cite{Graham_2021_ICCV,carion2020end}. For higher efficiency, FastViT~\cite{vasu2023fastvit} and DHVT~\cite{lu2022bridging} devise hybrid architectures to enhance computational and data efficiency, respectively. Another line of work explores the intrinsic relationships between the two types of operations~\cite{cordonnier2019relationship,chen2021x}. Among them, pioneer work~\cite{cordonnier2019relationship} shows that self-attention layers with the learnable quadratic position encoding and an abundant number of heads \emph{can express any convolutional layers}. Following~\cite{cordonnier2019relationship}, ConViT~\cite{d2021convit} includes soft inductive bias as partial of attention scores in self-attention layers and Transformed CNN~\cite{d2021transformed} learns to cast pre-trained convolutional layers into self-attention layers. Our approach also falls into the scope of the latter line of work. we propose a novel weight-sharing scheme showing that a subset of MSA operation parameters can express convolutional operations. In contrast to the previous works that are limited to ViTs with certain types of positional encodings and high head numbers, our weight-sharing scheme transcends these limitations and is applicable to a wide range of pre-trained ViTs. Furthermore, our weight-sharing scheme enables us to encode all candidate operations into MSA layers in a single-path search space to reduce the search cost.
\section{Method}

We start by introducing the proposed weight-sharing scheme between MSA and convolutional operations in Section~\ref{subsec:weight-sharing}. Then, we elaborate on our single-path vision transformer pruning approach that enjoys the benefits of the weight-sharing scheme in Section~\ref{subsec:SPViT}.

\subsection{Weight-sharing between MSA and Convolutional Operations} \label{subsec:weight-sharing}
The weight-sharing scheme refers to sharing a subset of parameters among different operations, which has been demonstrated to be effective by a breadth of NAS approaches~\cite{stamoulis2019single,guo2020single,cai2019once}, \eg, sharing weights among the candidate convolutional operations largely reduces the search cost~\cite{stamoulis2019single}. In this paper, we introduce a novel weight-sharing scheme between MSA and convolutional operations, which essentially helps us derive a single-path space that reduces the search cost and optimization difficulty.

\subsubsection{Revisiting Convolutional and MSA Layers}
Convolutional layers are the basic building blocks for CNNs. Let $\boldsymbol{X} \in \mathbb{R}^{n_w\times n_e\times c_{in}}$ and $\boldsymbol{W} \in \mathbb{R}^{k\times k\times c_{in}\times c_{out}}$ be the input features and the convolutional kernel with kernel size $k$, where $n_w$, $n_e$, $c_{in}$ and $c_{out}$ are the spatial width, height, input dimensions, and output dimensions, respectively. Standard convolutional layers aggregate the features within the local receptive field, which is defined in set ${\sDelta}:=[-\left \lfloor\frac{k}{2} \right \rfloor, ..., \left \lfloor\frac{k}{2} \right \rfloor] \times [-\left \lfloor\frac{k}{2} \right \rfloor, ..., \left \lfloor \frac{k}{2} \right \rfloor]$. Formally, let $\mathcal{S} := [1, ..., n_w]\times[1, ..., n_e]$ be the index set for the width and height for $\boldsymbol{X}$, the output for a standard convolutional layer at position $(i, j)\in\mathcal{S}$ can be expressed as
\begin{equation}
    \begin{aligned}
    {\rm Conv}(\boldsymbol{X})_{i, j, :} :=  \sum\limits_{(\delta_{1}, \delta_{2}) \in \sDelta} \boldsymbol{X}_{i+\delta_{1}, j+\delta_{2}, :} \boldsymbol{W}_{\delta_{1}, \delta_{2}, :, :}.
    \end{aligned}
\label{eq:convolution}
\end{equation}

Transformers take MSA layers as the main building blocks. However, instead of aggregating only neighboring features, MSA layers have larger receptive fields that cover the entire sequence. Considering the same input $\boldsymbol{X}$ as in Eq.~\eqref{eq:convolution} as a set of $c_{in}$ dimensional embeddings and letting $(l,m)\in\mathcal{S}$ be the index for any key feature embedding, the output for an MSA layer at position $(i, j)$ can be defined as
\begin{equation}
    \small
    \begin{aligned}
    {\rm MSA}(\boldsymbol{X})_{i, j, :}  & = \sum\limits_{h \in [n_h]} \sum\limits_{(l, m) \in \mathcal{S}} \operatorname{Att}(\boldsymbol{X})^h_{(i, j), (l, m)} \boldsymbol{V}_{l, m, h, :} \boldsymbol{W}^{\rm o}_{:,h,:},
    \end{aligned}
    \small
\label{eq:msa}
\end{equation}
where $\operatorname{Att}(\boldsymbol{X})^h =\operatorname{Softmax}\left(\boldsymbol{Q}_{:,:,h,:} \boldsymbol{K}_{:,:,h,:}^{T} / {\sqrt{c_{h}}}\right)$, $\boldsymbol{Q} := \boldsymbol{XW}^{\rm qry}$, $\boldsymbol{K} :=\boldsymbol{XW}^{\rm key}$, $\boldsymbol{V} :=\boldsymbol{XW}^{\rm val}$ and $n_h$ denotes the number of heads in MSA layers. Given the output dimension for the $h$-th head $c_h = c_{in} / n_h$, we define  $\boldsymbol{W}^{\rm val}$, $\boldsymbol{W}^{\rm qry}$ and $\boldsymbol{W}^{\rm key} \in \mathbb{R}^{c_{in}\times n_{h} \times c_{h}}$ as the corresponding learnable value, query and key linear projections. Therefore, $\operatorname{Att}(\boldsymbol{X})^h \in \mathbb{R}^{(n_wn_e)\times (n_wn_e)}$ denotes the attention map for the $h$-th head, and with the index of $(i,j)$ and $(l,m)$, ${\rm Att}(\boldsymbol{X})^h_{(i,j),(l,m)}$ becomes a scalar. With another learnable linear projection $\boldsymbol{W}^{\rm o} \in \mathbb{R}^{c_{h}\times n_{h}\times c_{in}}$, ${\rm MSA}(\boldsymbol{X})$ keeps the original dimension as $\boldsymbol{X}$. For simplicity, we ignore positional encodings~\cite{vaswani2017attention} in MSA layers and all the learnable bias terms accompanied with the learnable projections in both convolutional and MSA layers. We discuss that our weight-sharing scheme is not dependent on a certain type of positional encoding in Section~\ref{subsec:method_discussion}.

The computational complexity for MSA layers is $\mathcal{O}(n_en_wc_{in}^2 + (n_en_w)^2c_{in})$~\cite{hvt}. In this case, when modeling high-resolution feature maps that $n_en_w \gg c_{in}$, the second quadratic term for modeling global correlations dominates the computation. Whilst, in standard convolutional layers outputting the features maps with the same width and height, the computational complexity for convolutional layers remains as $\mathcal{O}(k^2n_en_wc_{in}c_{out})$.

\subsubsection{Weight-sharing Scheme}
\begin{figure}[t]
  \centering
  \includegraphics[width=\linewidth]{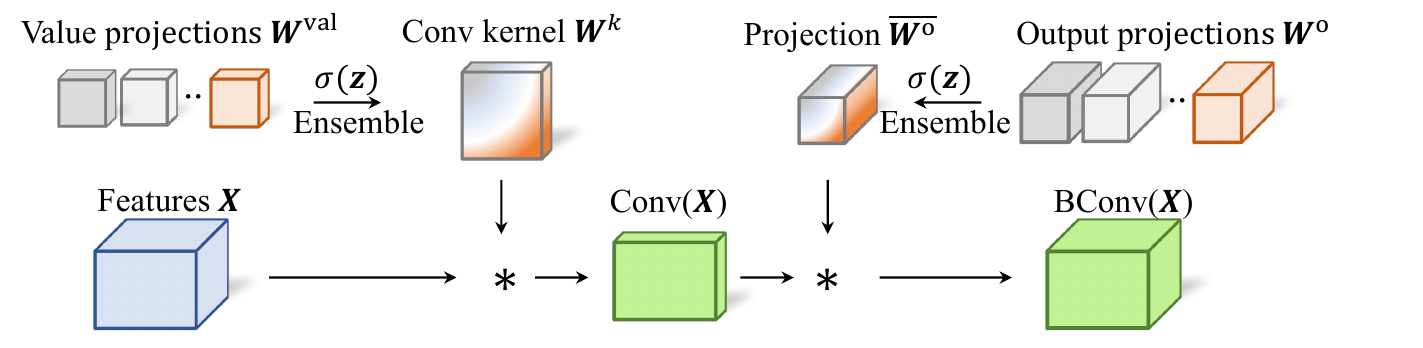}
  \caption{Overview of our weight-sharing scheme, which expresses the bottleneck convolution with MSA parameters. First, we restrict the value and output projections in the MSA layer to only process the features within local regions. Then, we learn to ensemble the value projections with a parameterized softmax $\sigma(\boldsymbol{z})$
  to derive the convolutional kernel. After convolution, we project the convolutional output to the original dimensionality by learning to ensemble the output projections and deriving the bottleneck convolution output.}
  \label{fig:derive_cnn}
\end{figure}

In this section, we demonstrate how a subset of MSA operation parameters can express bottleneck convolutional operations~\cite{he2016deep} when $c_{out}=c_{in}$, which is depicted in Figure~\ref{fig:derive_cnn}. In the following, we reveal the shared parameters for the two types of operations by gradually eliminating the parameters exclusive to the MSA operation.

First, since convolutional operations only process features within the local regions, we fix the attention score for the non-local correlations to $0$ and the local correlations to $1$ for each embedding when expressing convolutional operations with MSA parameters. Formally
\begin{equation}
    \small
    \begin{aligned}
    \operatorname{Att}(\boldsymbol{X})^h_{(i, j), (l, m)} =  \left\{\begin{array}{ll}1 & \text { if } (i - l, j - m) \in \sDelta \\ 0 & \text { otherwise }\end{array}\right..
    \end{aligned}
    \small
\label{eq:msa2conv0}
\end{equation}
By substituting back into Eq.~\eqref{eq:msa}, we have

\begin{equation}
    \small
    {\rm MSA}(\boldsymbol{X})_{i, j, :}^- = \sum\limits_{h \in [n_h]} \sum\limits_{(\delta_{1}, \delta_{2}) \in \sDelta} \boldsymbol{V}_{i+\delta_{1}, j+\delta_{2}, h, :} \boldsymbol{W}^{\rm o}_{:,h,:}.
    \small
\label{eq:msa2conv1}
\end{equation}
Here we use ${\rm MSA}(\boldsymbol{X})^-$ to represent outputs with the locality restriction in Eq.~\eqref{eq:msa2conv0}.
In Eq.~\eqref{eq:msa2conv1}, we force the heads of an MSA operation to attend to the positions within the local window of size $k\times k$ centered at $(i, j)$.

Next, we show that we can further profile MSA operation parameters into a convolutional kernel. Analogous to~\cite{cordonnier2019relationship}, one way is defining a bijective mapping $\boldsymbol{f}:\left[n_h'\right] \rightarrow \sDelta$ that assigns heads in a selected head subset $\left[n_h'\right]$ to specific positions within the local windows that $[n_h'] \subseteq [n_h]$. We can then substitute $h$ with $\boldsymbol{f}(\delta_{1}, \delta_{2})$ in Eq.~\eqref{eq:msa2conv1}, \ie,
\begin{equation}
    \small
    \begin{aligned}
    {\rm MSA}(\boldsymbol{X})_{i, j, :}^- & = \sum\limits_{(\delta_{1}, \delta_{2})\in \sDelta} \boldsymbol{V}_{i+\delta_{1}, j+\delta_{2}, \boldsymbol{f}(\delta_{1}, \delta_{2}), :} \boldsymbol{W}^{\rm o}_{:,\boldsymbol{f}(\delta_{1}, \delta_{2}),:}.
    \end{aligned}
    \small
\label{eq:msa2conv2}
\end{equation}
Here in Eq.~\eqref{eq:msa2conv2}, each head attends to a certain position within the local window. Nevertheless, many ViT variants have $n_h < 9$, \eg, DeiT-Ti and DeiT-S~\cite{touvron2021training}, which would restrict the weight-sharing scheme to only the large architectures. Also, in MSA layers, different heads tend to attend to different areas as visualized in~\cite{cordonnier2019relationship,pan2021less}. Hence, it is non-trivial to define the subset $[n_h']$ and the mapping function $\boldsymbol{f}$ which identifies the most suitable heads for the local positions. To allow the weight-sharing scheme to be applicable for generic ViT architectures as
well as ease the difficulty for defining $[n_h']$ and $\boldsymbol{f}$, we employ learnable parameters $\boldsymbol{z}\in \mathbb{R}^{n_h\times k\times k}$ followed by a softmax function $\sigma(\cdot)$ to first ensemble the MSA heads into the convolutional kernel positions:
\begin{equation}
\begin{small}
\begin{aligned}
    \!{\rm MSA}(\boldsymbol{X})_{i, j, :}^-\!&=\!\sum\limits_{h \in [n_h]}\!\sum\limits_{(\delta_{1}, \delta_{2})\in \sDelta} \!\sigma(\mathbf{z})_{h, \delta_1, \delta_2}\!\boldsymbol{V}_{i+\delta_{1}, j+\delta_{2}, h, :}\! \boldsymbol{W}^{\rm o}_{:,h,:},
\end{aligned}
\label{eq:msa2conv3}
\end{small}
\end{equation}
where $\sigma(\mathbf{z})_{h, \delta_1, \delta_2} =e^{\mathbf{z}_{h, \delta_1, \delta_2}} / \sum_{k=1}^{n_h} e^{\mathbf{z}_{k, \delta_1, \delta_2}}.$
In this case, the ensembles of MSA heads with scaling factor $\sigma(\mathbf{z})_{h, \delta_1, \delta_2}$ learn to attend to pre-defined positions within the local $k\times k$ windows. We make sure the ensembles of heads remain the original scale by employing the softmax function on $\boldsymbol{z}$. 

Next, according to~\cite{cordonnier2019relationship}, it is possible to profile $\boldsymbol{W}^{\rm val}_{:,h,:}\boldsymbol{W}^{\rm o}_{:,h,:}$ into a convolutional kernel from Eq.~\eqref{eq:msa2conv3} by substituting $\boldsymbol{V}$ with $\boldsymbol{XW}^{\rm val}$ and applying the associative law:
\begin{equation}
\begin{small}
    \begin{aligned}
    \wh{{\rm Conv}}(\boldsymbol{X})_{i,j,:} &= \sum\limits_{(\delta_{1}, \delta_{2})\in \sDelta}\boldsymbol{X}_{i+\delta_{1}, j+\delta_{2}, :} \boldsymbol{\wh{W}}_{\delta_{1}, \delta_{2},:,:},
    \end{aligned}
\label{eq:msa2conv4}
\end{small}
\end{equation}
where $\boldsymbol{\wh{W}}_{\delta_{1}, \delta_{2},:,:}\!:=\!\sum\limits_{h \in [n_h]}\sigma(\mathbf{z})_{h, \delta_1, \delta_2}\left(\boldsymbol{W}^{\rm val}_{:,h,:}\boldsymbol{W}^{\rm o}_{:,h,:}\right).$ Instead, for the complexity consideration, we choose to form a bottleneck convolution operation~\cite{he2016deep} and modify Eq.~\eqref{eq:msa2conv3} as
\begin{small}
    \begin{align}
    &{\rm BConv}(\boldsymbol{X})_{i, j, :} 
    \notag\\ 
    =& \Big(\!\sum\limits_{h \in [n_h]}\!\sum\limits_{(\delta_{1}, \delta_{2})\in \sDelta} \!\sigma(\mathbf{z})_{h, \delta_1, \delta_2}\!\boldsymbol{V}_{i+\delta_{1}, j+\delta_{2}, h, :}\! \Big) \overline{\boldsymbol{W}^{\rm o}} \label{eq:msa2conv5_1} \\
    =& \Big(\sum\limits_{(\delta_{1}, \delta_{2})\in \sDelta}\boldsymbol{X}_{i+\delta_{1}, j+\delta_{2}, :} \boldsymbol{W}_{\delta_{1}, \delta_{2},:,:} \Big) \overline{\boldsymbol{W}^{\rm o}} \label{eq:msa2conv5_2} \\
    \Rightarrow& {\rm Conv}(\boldsymbol{X})_{i,j,:}\overline{\boldsymbol{W}^{\rm o}} \label{eq:msa2conv5_3},
    \end{align}
\end{small}
where
\begin{equation}
\begin{small}
    \begin{aligned}
    \left\{\begin{array}{ll}\boldsymbol{W}_{\delta_{1}, \delta_{2},:,:} := \sum\limits_{h \in [n_h]}\sigma(\mathbf{z})_{h, \delta_1, \delta_2}\boldsymbol{W}^{\rm val}_{:,h,:} \\ \overline{\boldsymbol{W}^{\rm o}} := \sum\limits_{h \in [n_h]}\sum\limits_{\delta_{1}, \delta_{2}\in \sDelta}\sigma(\mathbf{z})_{h, \delta_1, \delta_2}\boldsymbol{W}^{\rm o}_{:,h,:} \end{array}\right..
    \end{aligned}
\label{eq:msa2conv6}
\end{small}
\end{equation}
Note that in this form, the convolutional operation output dimension $c_{out} = c_{h}$ and we project ${\rm Conv}(\boldsymbol{X})\in \mathbb{R}^{n_w\times n_e\times c_{h}}$ back to $\mathbb{R}^{n_w\times n_e\times c_{in}}$ via $\overline{\boldsymbol{W}^{\rm o}}$. One can easily find that both ${\rm Rank}(\wh{{\rm Conv}}(\boldsymbol{X}))$ in Eq.~\eqref{eq:msa2conv4} and ${\rm Rank}({\rm Conv}(\boldsymbol{X})\overline{\boldsymbol{W}^{\rm o}})$ in Eq.~\eqref{eq:msa2conv5_3}
are not greater than $c_h$, the dimension of heads. However, the computational complexity for the former is $\mathcal{O}(k^2n_wn_ec_{in}^2)$ and for the latter is $\mathcal{O}(k^2n_wn_ec_{in}c_h)$. Since $c_h \leq c_{in}$, we choose to employ the bottleneck convolutions for better efficiency.

We further add batch normalization ($\rm{BN}$)~\cite{nair2010rectified} and $\rm{ReLU}$ non-linearity~\cite{ioffe2015batch} and get the bottleneck convolutional operations expressed by $\boldsymbol{W}^{\rm val}$ and $\boldsymbol{W}^{\rm o}$:
\begin{equation}
\small
\begin{aligned}
{\rm BConv}(\boldsymbol{X})_{i,j,:} &= {\rm BN}({\rm ReLU}({\rm Conv}\left(\boldsymbol{X}\right)_{i,j,:}))
\overline{\boldsymbol{W}_{\rm o}}.
\end{aligned}
\small
\label{eq:msa2conv7}
\end{equation}

\subsubsection{Discussions}\label{subsec:method_discussion}

The proposed weight-sharing scheme encourages the attention heads of an MSA operation to model the global regions and learn ensembles to process the local region features simultaneously. The intuition for this approach mainly comes from two aspects. Firstly, the local regions are indeed part of the global regions, and thus the ensemble of MSA heads naturally has the capacity for handling local regions. Secondly, since some attention heads in ViT encoders often attend to the local regions around the query pixels~\cite{cordonnier2019relationship,raghu2021vision}, expressing the behaviors for convolutional operations with these heads is likely to get us reasonable outputs. However, considering the optimal places to add locality might vary for different ViT models, we thus introduce SPViT that automatically learns to choose the optimal operations in Section~\ref{subsec:SPViT}.

In a nutshell, there are two main differences when comparing the proposed weight-sharing scheme with~\cite{cordonnier2019relationship}. First, the weight-sharing scheme has broader applications on ViT models without the conditions of using the relative positional encoding (quadratic encoding) and $n_h \geq k^2$ as required in~\cite{cordonnier2019relationship}. The reason is that~\cite{cordonnier2019relationship} learns the attention head subset $[n_h']$ and the bijective mapping $\boldsymbol{f}$ in Eq.~\eqref{eq:msa2conv2} between heads and the convolutional kernel positions with quadratic encoding. While we simplify this process by forming the convolutional kernels as ensembles of attention heads with $\sigma(\boldsymbol{z})$. Second, we propose to use a bottleneck convolutional operation with the same rank upper bound but higher efficiency compared to the standard convolutional operation in~\cite{cordonnier2019relationship}.

\begin{figure*}[t]
  \centering
  \includegraphics[width=\linewidth]{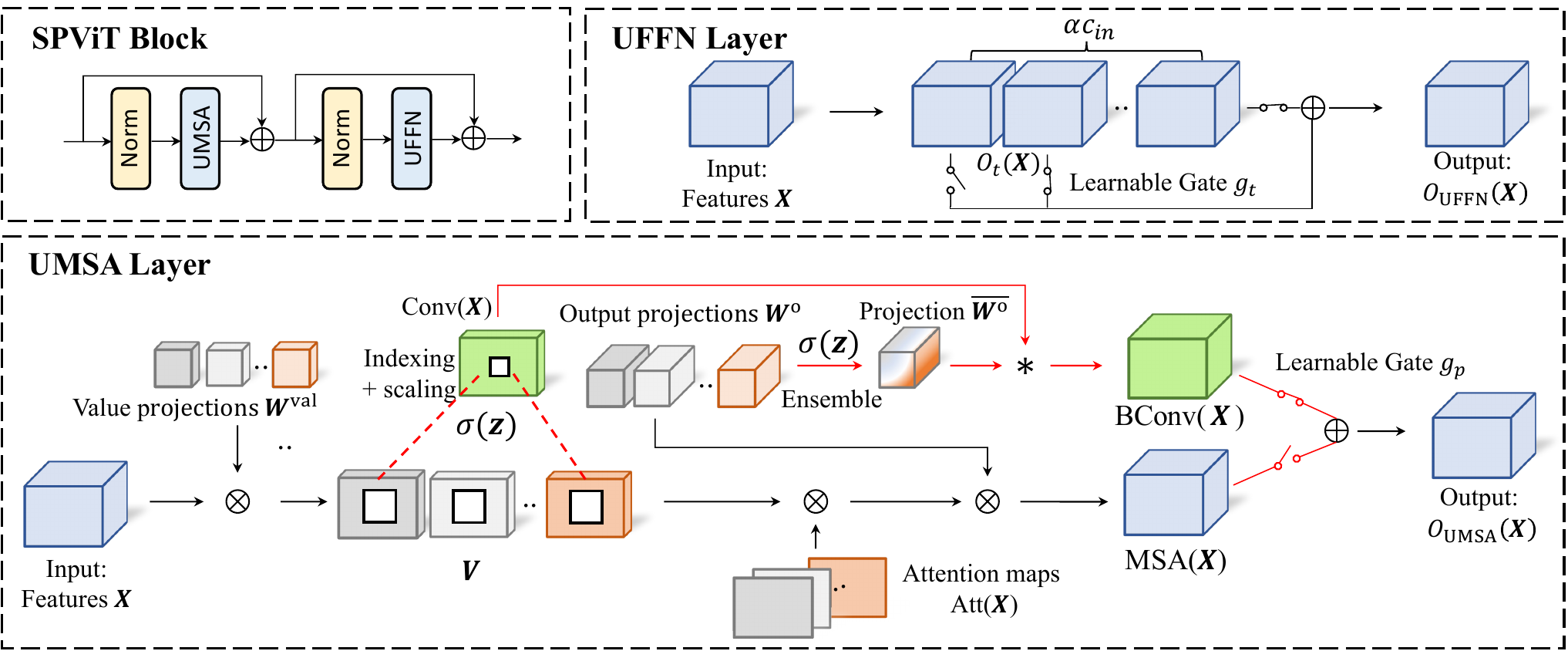}
\caption{The overview of our SPViT during the search. \textbf{SPViT block (top left)}: We replace the original MSA and FFN layers with our UMSA and UFFN layers to search for a compact architecture, while keeping the other components intact. \textbf{UFFN layer (top right)}: We search for important FFN hidden dimensions. Given the input features $\boldsymbol{X}$, we employ learnable binary gates for each of the $\alpha c_{in}$ transformed features $O_t(\boldsymbol{X})$, where $\alpha$ and $c_{in}$ represent the MLP expansion ratio and input dimensions. We optimize the learnable binary gates to search pruned FFN structures. \textbf{UMSA layer (bottom)}: We search for MSA or convolutional layers in single path. In addition to the standard MSA operation (lower branch as described by Eq.~(\ref{eq:msa})), we get the output of bottleneck convolutional layers (upper branch). The additional steps are in red lines where we index and scale the values $\boldsymbol{V}$ with the scaling factors $\sigma(\mathbf{z})$ to get the convolution output ${\rm Conv}{(\boldsymbol{X})}$, and then derive the bottleneck convolution output ${\rm BConv}(\boldsymbol{X})$ as described in Eq.~(\ref{eq:msa2conv7}). For simplicity, we only show deriving the ${\rm Conv}{(\boldsymbol{X})}$ of one kernel size. In this way, we get both the MSA output and the bottleneck convolution outputs simultaneously and finally optimize the learnable gates to select operations.}
\label{fig:main}
\end{figure*}

\subsection{Single-path Vision Transformer Pruning}\label{subsec:SPViT}

We introduce our SPViT method that prunes both MSA and FFN layers with NAS. To do so, we replace the original MSA and FFN layers with our unified MSA (UMSA in Section~\ref{subsec:umsa}) and unified FFN (UFFN in Section~\ref{subsec:uffn}) layers, as depicted in Figure~\ref{fig:main} top left. Both layers encode the architecture configuration choices with learnable binary gates to form a unified search space. We optimize these binary gates and model weights with an overall objective (Section~\ref{subsec:objective}) to explore a wide range of efficient architectures. Finally, we fine-tune the searched architectures and get the compact pruned ViTs (Section~\ref{subsec:ft}).

\subsubsection{Searching for MSA or Convolutional layers} \label{subsec:umsa}
We first search for the optimal places to prune MSA layers into convolutional layers in ViTs with our UMSA layers as depicted in Figure~\ref{fig:main} bottom. Specifically, with our weight-sharing scheme in Section~\ref{subsec:weight-sharing}, we devise a single-path search space that directly get the convolution output ${\rm Conv}(\boldsymbol{X})$ for pre-defined kernels $k\in \mathcal{K}$ by indexing and scaling $\boldsymbol{V}$ according to Eq.~\eqref{eq:msa2conv5_1}, without the need to keep each candidate convolutional operation as a separate path. Next, we get the bottleneck convolution output ${\rm BConv}(\boldsymbol{X})$ with Eq.~\eqref{eq:msa2conv7}.
Thanks to the single-path search space, we derive the outputs for both bottleneck convolutions and MSA with low computational cost. By the order of computational complexity from low to high, we can formulate the bottleneck convolutions and MSA into an ordered set $\mathcal{P}$, where $|\mathcal{P}|$ = $|\mathcal{K}| + 1$.

Next, following~\cite{liu2021elastic,li2020differentiable}, we define a series of binary gates to encode the operation choices. For any operation with index $p\in[|\mathcal{P}|]$, gate $g_p$ is sampled from a Bernoulli distribution with a learnable gate parameter $\theta_{p}$ which can be automatically optimized. Specifically, we use ${\rm Sigmoid}(\theta_p)$ to determine the probability of choosing the $p$-th operation:
\begin{equation}
    g_p\sim {\rm Ber}({\rm Sigmoid}(\theta_{p})).
\label{eq:gate1}
\end{equation}
For efficiency consideration, we choose only one operation per layer. In this way, we enforce each UMSA layer to have at most one open gate by replacing $g$ with $\hat{g}$: $\hat{g_p} = g_p \prod_{q=p+1}^{|\mathcal{P}|} \left (1 - g_q \right )$. It is essentially stated that once a more complex gate is selected ($g_q=1$), all less complex gates will be closed. The output for the UMSA layer hence can be expressed as
\begin{equation}
    \small
    \begin{aligned}
    O_{\rm UMSA}(\boldsymbol{X}) = & \sum_{p=1}^{|\mathcal{P}|} \hat{g_p} \cdot O_{p}(\boldsymbol{X}),
    \end{aligned}
    \small
\label{eq:gate3}
\end{equation}
where $O_{p}(\boldsymbol{X})$ is the output for the $p$-th operation.
Each UMSA layer is then followed by a residual connection and layer normalization~\cite{ba2016layer} as ViTs do. The skip-connection operation is adopted when all gates are closed. Hence, our UMSA search space includes three types of candidate operations: skip-connections, bottleneck convolutions with different candidate kernel sizes, and MSA.

\subsubsection{Searching for Fine-grained MLP Expansion Ratios}
\label{subsec:uffn}
FFN layers also consume a large number of FLOPs. In this work, we jointly include MSA and FFN pruning configurations into a unified search space and flexibly adjust the MSA-FFN proportions for each specific dense model. To do so, we prune the unimportant FFN hidden dimensions to search for fine-grained MLP expansion ratios.

MLP expansion ratios control the number of hidden dimensions for the FFN layers. Previous works~\cite{chen2021autoformer,liao2021searching,chen2021glit} define the search space for the MLP expansion ratios as a list of coarse-grained values, \eg, $[3,3.5,4]$ in~\cite{chen2021autoformer}. Instead, given the MLP expansion ratio $\alpha$ defined by the pre-trained model,
we propose a simple yet effective approach to search for fine-grained MLP expansion ratios $\alpha'$ that $0\leq \alpha' \leq \alpha$ for each FFN layer via our UFFN layer. The overview of our UFFN layer is depicted in Figure~\ref{fig:main} top right. Given input features $\boldsymbol{X}$, the output for a FFN layer can be expressed as $O_{\rm FFN}(\boldsymbol{X}) = \sum\limits_{t\in [\alpha c_{in}]} O_{t}(\boldsymbol{X})$, where $O_{t}(\boldsymbol{X}) = {\rm GeLU}\left (\boldsymbol{XW}^{\rm fc1}_{:, t} \right )\boldsymbol{W}^{\rm fc2}_{t, :}.$ Here, we use $\alpha$ and $\alpha c_{in}$ to denote the pre-defined MLP expansion ratio and the hidden dimension for FFN layers. $\boldsymbol{W}^{\rm fc1}\in \mathbb{R}^{c_{in}\times \alpha c_{in}}$ and $\boldsymbol{W}^{\rm fc2}\in \mathbb{R}^{\alpha c_{in}\times c_{in}} $ are learnable fully-connected projections where the former maps the channel dimension of $\boldsymbol{X}$ to $\alpha c_{in}$ and the latter projects the channel dimension back to $c_{in}$. $\rm{GeLU}$~\cite{hendrycks2016gaussian} activation is added between the two learnable fully-connected projections.

Like Eq.~\eqref{eq:gate1}, we encode FFN hidden dimension choices with binary gates $g_t$. By applying the binary gates to each hidden dimension during the search, we define the UFFN layer output as
\begin{equation}
\small
O_{\rm UFFN}(\boldsymbol{X}) = \sum\limits_{t\in [\alpha c_{in}]} g_t \cdot O_{t}(\boldsymbol{X}).
\small
\label{eq:search_ffn_2}
\end{equation}
This allows us to identify the most important hidden dimensions thereby searching for FFN layers with fine-grained MLP expansion ratios.

\subsubsection{Searching Objective}\label{subsec:objective}

To get the architectures with desired efficiency constraints, we optimize the network with an auxiliary computational complexity loss. Specifically, we define a look-up table containing the computational complexities for the candidate operations and modules. The computational cost $\cL_{\rm comp}$ is defined as $\cL_{\rm comp}  =  ({\rm F}(\boldsymbol{X}) - {\hat{\rm F}(\boldsymbol{X})})^2$,
where ${\rm F}(\boldsymbol{X})$ is the current network computational complexity and  $\hat{\rm F}(X)$ is the target one.
Finally, the overall searching objective is defined as $\cL = \cL_{\rm CE} + \lambda_{\rm comp}\cL_{\rm comp}$, where $\lambda_{\rm comp}$ is the trade-off hyper-parameter
balancing $\cL_{\rm comp}$ and the cross-entropy loss $\cL_{\rm CE}$. Benefiting from the weight-sharing scheme between MSA and convolutional operations, the search cost is largely reduced, as shown in Section~\ref{subsec:abl_single_path}.

With our unified search space and the global loss, we explore a wide range of efficient architectures and flexibly adjust the MSA-FFN proportions for each specific dense model (Sections~\ref{subsec:pruning_proportion}) and make intriguing observations on the preferred compact architectures (Section~\ref{subsec:observation}).

\subsubsection{Fine-tuning} \label{subsec:ft}
During fine-tuning, we make the previous stochastic binary gates in Eqs.~\eqref{eq:gate1}-\eqref{eq:search_ffn_2} deterministic, \ie, $\hat{g_p}' = \mathbbm{1}({\rm Sigmoid}(\theta_p) \geq 0.5)$.
Same as the search, we obtain $\hat{g_p}'$ and select only at most one operation in UMSA layers during fine-tuning. The output can be expressed as
\begin{equation}
    \small
    \begin{aligned}
    O'_{\rm UMSA}(\boldsymbol{X}) =
    \left\{\begin{array}{ll} 
    {\rm MSA}(\boldsymbol{X}) & \text { if } \hat{g_p}'=1 \quad \text { and } \quad p=|\mathcal{P}| \\
    {\rm BConv}'(\boldsymbol{X}) & \text { if } \hat{g_p}'=1 \quad \text { and } \quad p<|\mathcal{P}| \\
    0 & \text { otherwise}
    \end{array}\right..
    \end{aligned}
    \small
\label{eq:ft_2}
\end{equation}
If the selected operation is bottleneck convolution, we can profile the pre-trained $\boldsymbol{W}^{\rm val}$ and $\boldsymbol{W}^{\rm o}$ into convolutional kernels and directly perform convolutions as described in Eq.~\eqref{eq:msa2conv5_3} and depicted in Figure~\ref{fig:derive_cnn}. The other MSA parameters are slimmed. Combining with Eq.~\eqref{eq:msa2conv7}, we can define the bottleneck convolutional layer during fine-tuning as ${\rm BConv}'(\boldsymbol{X}) = {\rm BN}({\rm ReLU}(\boldsymbol{X}\ast \boldsymbol{W})\overline{\boldsymbol{W}_{\rm o}},$ where $\boldsymbol{W}={\rm Arrange}(\boldsymbol{W}_{1, 1,:,:}, ..., \boldsymbol{W}_{k, k,:,:})$. Similarly, for UFFN layers, we can automatically find a subset of hidden dimensions $\mathcal{T} \subseteq [\alpha c_{in}]$ indicating the selected hidden dimensions. Therefore, $\alpha'$ is $|\mathcal{T}|/c_{in}$ and the UFFN layer outputs during fine-tuning can be derived by $O'_{\rm UFFN}(X) = \sum\limits_{t\in \mathcal{T}} O_t(\boldsymbol{X})$.

We further incorporate our SPViT with knowledge distillation~\cite{hinton2015distilling}, which is a commonly utilized technology to compensate for the accuracy loss in the model compression scenario~\cite{yang2021nvit, yu2022unified}. Specifically, we let the pruned network learn from the hard labels predicted by the teacher network, akin to~\cite{hinton2015distilling}. As our SPViT introduces convolutional inductive bias to the ViT models, it facilitates guidance from a CNN teacher (see Section~\ref{subsec:abl_each}). Therefore, the final objective during fine-tuning is:
\begin{equation}
\small
\begin{aligned}
\cL = \cL_{\rm CE} + \lambda_{\rm dist}\cL_{\rm dist},
\end{aligned}
\small
\label{eq:ft_objective}
\end{equation}
where $\cL_{\rm dist}$ is the hard-label distillation loss between the CNN teacher and the pruned ViT student and $\lambda_{\rm dist}$ is a balancing trade-off hyper-parameter. Note that although we employ a CNN teacher similar to DeiT~\cite{touvron2021training}, our SPViT does not require an additional distillation token as DeiT does, thereby introducing no computational overhead during inference.
\section{Experiments}

\begin{table*}[t!]
\centering
\caption{Comparison with the competitor pruning methods on ImageNet-1k for pruning DeiT~\cite{touvron2021training} and Swin~\cite{swin}. * indicates sparse training from scratch methods. $\S$ indicates fine-tuning with knowledge distillation.
}
\resizebox{0.9\textwidth}{!}{%
\begin{tabular}{c|ccccc}
Models & FLOPs (G) & Top-1 Acc. (\%) & Top-1 Acc. loss (\%) & Top-5 Acc. (\%) & Params (M) \\ \hline
DeiT-Ti~\cite{touvron2021training} & 1.3 & 72.2 & - & 91.1 & 5.7  \\
SSP-DeiT-Ti*~\cite{michel2019sixteen,bartoldson2019generalization} & 1.0 & 68.6 & 3.6 & - & 4.2  \\
S$^2$ViTE-DeiT-Ti*~\cite{chen2021chasing}  & 1.0 & 70.1 & 2.1 & - & 4.2 \\
UVC-DeiT-Ti$\S$~\cite{yu2022unified} & 0.7 & 71.8 & 0.4 & - & -\\
SPViT-DeiT-Ti & 1.0 & 70.7 & 1.5 & 90.3 & 
4.8\\
SPViT-DeiT-Ti$\S$ & 1.0 & \textbf{73.2} & \textbf{-1.0} & \textbf{91.4} & 4.8\\
\hline
DeiT-S~\cite{touvron2021training} & 4.6 & 79.9 & - & 95.0 & 22.1 \\
SSP-DeiT-S*~\cite{michel2019sixteen,bartoldson2019generalization} & 3.2 & 77.7 & 2.2 & - & 14.6 \\
S$^2$ViTE-DeiT-S*~\cite{chen2021chasing}  & 3.2 & 79.2 & 0.7 & - & 14.6\\
DynamicViT-DeiT-S~\cite{rao2021dynamicvit} & 3.0 & 79.3 & 0.6 & 94.7 & 22.8 \\
EViT-DeiT-S~\cite{liang2022not} & 3.0 & 79.5 & 0.4 & 94.8 & 22.1 \\
eTPS-DeiT-S~\cite{wei2023joint} & 3.0 & 79.7 & 0.2 & - & 22.1 \\
dTPS-DeiT-S~\cite{wei2023joint} & 3.0 & 80.1 & -0.2 & - & 22.8 \\
MDC-DeiT-S~\cite{hou2021multi} & 2.9 & 79.9 & 0.0 & - & -\\
ToMe-DeiT-S~\cite{bolya2022token} & 2.7 & 79.4 & 0.5 & - & 22.1 \\
UVC-DeiT-S$\S$~\cite{yu2022unified} & 2.7 & 79.4 & 0.5 & - & -\\
SPViT-DeiT-S & 3.3 & 78.3 & 1.6 & 94.3 & 15.9 \\
SPViT-DeiT-S$\S$ & 3.3 & \textbf{80.3} & \textbf{-0.4} & \textbf{95.1} & 15.9 \\\hline
DeiT-B~\cite{touvron2021training} & 17.5 & 81.8 & - & 95.6 & 86.4 \\
VTP-DeiT-B~\cite{zhu2021visual} & 13.8 & 81.3 & 0.5 & 95.3 & 67.3 \\
SSP-DeiT-B*~\cite{michel2019sixteen,bartoldson2019generalization} & 11.7 & 80.1 & 1.7 & - & 56.8 
\\
IA-RED$^2$-DeiT-B~\cite{pan2021ia} & 11.8 & 81.3 & 0.5 & - & 86.4 \\
S$^2$ViTE-DeiT-B*~\cite{chen2021chasing}  & 11.7 & 82.2 & -0.4 & - & 56.8\\
EViT-DeiT-B~\cite{liang2022not} & 11.6 & 82.1 & -0.3 & 95.6 & 86.4 \\
eTPS-DeiT-B~\cite{wei2023joint} & 11.4 & 81.1 & 0.7 & - & 86.4 \\
dTPS-DeiT-B~\cite{wei2023joint} & 11.4 & 81.2 & 0.6 & - & 87.0 \\
DynamicViT-DeiT-B~\cite{rao2021dynamicvit} & 11.4 & 81.4 & 0.4 & 95.5 & 87.1 \\
MDC-DeiT-B~\cite{hou2021multi} & 11.2 & 82.3 & -0.5 & - & -\\
VTP-DeiT-B~\cite{zhu2021visual}  & 10.0  & 80.7 & 1.1 & - & 48.0 \\
UVC-DeiT-B$\S$~\cite{yu2022unified}  & 8.0 & 80.6 & 1.2 & - & -
\\
SPViT-DeiT-B  & 8.4  & 81.5 & 0.3 & 95.7 & 41.6 \\
SPViT-DeiT-B$\S$ & 8.4  & \textbf{82.4} & \textbf{-0.6} & \textbf{96.1}  & 41.6 \\\hline
Swin-Ti~\cite{swin} & 4.5 & 81.2 & - & 95.5 & 28.3 \\
STEP-Swin-Ti~\cite{li2021differentiable} & 3.5  & 77.2 & 4.0 & 93.6 & 23.6\\
SPViT-Swin-Ti & 3.4 & 80.1 & 1.1 & 95.0 & 25.8 \\
SPViT-Swin-Ti$\S$ & 3.4 & \textbf{81.0} & \textbf{0.2} & \textbf{95.4} & 25.8 \\
\hline
Swin-S~\cite{swin} & 8.7 & 83.0 & - & 96.2 & 49.6 \\
STEP-Swin-S~\cite{li2021differentiable} & 6.3 & 79.6 & 3.4 & 94.7 & 36.9 \\
SPViT-Swin-S & 6.1 & 82.4 & 0.6 & 96.0 & 38.9 \\
SPViT-Swin-S$\S$ & 6.1 & \textbf{83.0} & \textbf{0.0} & \textbf{96.4} & 38.9 \\
\end{tabular}%
}
\label{tab:swin}
\end{table*}

\begin{table}[t!]
\centering
\caption{Effect of our single-path search with SPViT-DeiT-Ti and SPViT-DeiT-S on ImageNet-1k.
``SPViT-Model w/MP'' denotes our pruning method except changing single-path to multi-path.}
\vspace{-0.8em}
\resizebox{0.48\textwidth}{!}{%
    \begin{tabular}{c|cccc}
    Model & \begin{tabular}[c]{@{}c@{}}Top-1 Acc. \\ (\%)\end{tabular} & \begin{tabular}[c]{@{}c@{}} Params \\  (M)\end{tabular}  & \begin{tabular}[c]{@{}c@{}}FLOPs \\ (G)\end{tabular}   & \begin{tabular}[c]{@{}c@{}}Search Cost\\ (GPU hours)\end{tabular} \\ \shline
    SPViT-DeiT-Ti w/ MP & 70.5 & 5.0 & 1.0 & 18.1 \\
    SPViT-DeiT-Ti & \textbf{70.7} & \textbf{4.8} & 1.0 & \textbf{8.6} \\  \hline
    SPViT-DeiT-S w/ MP & 78.0 & 16.5 & 3.3 & 28.9 \\
    SPViT-DeiT-S & \textbf{78.3} & \textbf{15.9} & 3.3 & \textbf{12.0} \\
    \end{tabular}
}
\label{tab:abl_single}
\end{table}

\begin{table*}[tb!]
\centering
\caption{Ablation study with SPViT-DeiT-Ti and SPViT-DeiT-B on ImageNet-1k.}
\vspace{-0.8em}
\resizebox{\textwidth}{!}{
\begin{tabular}{ccc|ccc|ccc} 
&&& \multicolumn{3}{c|}{\bf{SPViT-DeiT-Ti}} & \multicolumn{3}{c}{\bf{SPViT-DeiT-B}}\\
Prune MSA & Prune FFN &  Distillation & Top-1 Acc. (\%) & FLOPs (G)  & Params (M) & Top-1 Acc. (\%) & FLOPs (G)  & Params (M) \\
\hline
& & & 72.2 & 1.3 & 5.7 & 81.8 & 17.5 & 86.4 \\
\hline
\checkmark & & & 70.7 & 1.0 & 5.0 & 81.4 & 13.3 & 69.9 \\
& \checkmark & & 70.6 & 1.0 & 4.4 & 81.3 & 13.3 & 65.1 \\
\checkmark  & \checkmark &  & 70.7 & 1.0 & 4.8 & 81.5 & 8.4 & 41.6 \\
\checkmark & \checkmark &  \checkmark & 73.2 & 1.0 & 4.8 & 82.4 & 8.4 & 41.6 \\
\end{tabular}
}
\label{tab:ablation}
\end{table*}

\label{subsec:experiments}
In this section, we validate the effectiveness of our SPViT by pruning different ViT models. We conduct experiments on two series of ViT models: DeiT~\cite{touvron2021training} and Swin~\cite{swin}, which are representative standard ViT and hierarchical ViT~\cite{wang2021pyramid,hvt} models. In the following, we first compare with the baseline pruning methods in Section~\ref{subsec:compare}, then ablate the important components of SPViT in Section~\ref{subsec:ablation}, and finally analyze some observed patterns when varying the target computational complexity in Section~\ref{subsec:observation}.

\subsection{Experimental Setup}

\subsubsection{Implementation Details}
We perform both searching and fine-tuning from the pre-trained models. Unless specified, the training details for our SPViT align with those introduced in DeiT~\cite{touvron2021training} and Swin~\cite{swin}. During the search, we set the initial learning rates for the gate parameters $\theta$ and other network parameters as $1\times10^{-3}$ and $1\times10^{-4}$, respectively. The initialized value for the gate parameters $\theta$ is set to 1.5 to enable a high initial probability for choosing MSA operations. In UMSA layers, we include the most popular $1\times 1$ and $3\times 3$ bottleneck convolutions in our search space. We grid search and set different hyper-parameters $\lambda_{\rm comp}$ when pruning different models, according to their redundancy. We set $\lambda_{\rm dist}$ in Eq.~\eqref{eq:ft_objective} to 1.0. The network architectures converge at around 10 epochs on ImageNet-1k, and then, we follow~\cite{yu2022unified,liu2021lbs} to fine-tune %
    from the searched architectures with a learning rate of $1\times10^{-4}$. For experiments on ImageNet-1k~\cite{russakovsky2015imagenet}, we search and fine-tune our models with a total batch size of 1,024 on 8 NVIDIA V100 GPUs, except that SPViT-DeiT-Ti is trained on 4 NVIDIA V100 GPUs. We also evaluate our models with the same hardware specifications.

\subsubsection{Compared Methods}
When pruning DeiT~\cite{touvron2021training}, we compare with previous state-of-the-art ViT module pruning methods, including VTP~\cite{zhu2021visual},  S$^2$ViTE~\cite{chen2021chasing}, UVC~\cite{yu2022unified}, and MDC~\cite{hou2021multi}. We refer the readers to Section~\ref{subsec:related_work_pruning} for more details. We also compare with Salience-based Structured Pruning (SSP)~\cite{michel2019sixteen,bartoldson2019generalization}, which removes sub-modules by leveraging their weights, activations, and gradient information (implementation borrowed from~\cite{chen2021chasing}). In addition, we compare with representative token compression methods. For token pruning methods~\cite{li2021dynamic, pan2021ia, liang2022not}, DynamicViT~\cite{li2021dynamic} and IA-RED$^2$~\cite{pan2021ia} progressively estimate the uninformative tokens to prune with distinct prediction modules, while EViT~\cite{liang2022not} prunes the tokens with lowest attention scores. For token merging methods~\cite{bolya2022token, wei2023joint}, ToMe~\cite{bolya2022token} merges the most alike tokens while eTPS~\cite{wei2023joint} and dTPS~\cite{wei2023joint} merge the unimportant tokens to the important ones. Since we are the pioneering work that prunes Swin~\cite{swin}, we tailor the Straight-Through Estimator Pruning method (STEP)~\cite{li2021differentiable} as the comparing baseline. STEP learns the importance scores for attention heads of MSA layers and hidden dimensions of FFN layers with the learnable gate parameters optimized by the Straight-Through Estimator~\cite{bengio2013estimating}, which is related to our SPViT. The top-k attention heads and hidden dimensions with the highest importance scores are kept.

\subsection{Main Results}\label{subsec:compare}
We investigate the effectiveness of our method by comparing SPViT with baseline methods in Table~\ref{tab:swin}. 

\emph{\textbf{First, SPViT variants perform favorably against the module pruning methods.}}
For instance, our SPViT-DeiT-B compresses the DeiT-B to 48.0\% of the original FLOPs while incurring only 0.3\% Top-1 accuracy drop, outperforming the SOTA method UVC-DeiT-B$\S$ by 0.9\% Top-1 accuracy. We conjecture that the hybrid architectures searched by SPViT variants can learn more discriminative feature representations, thereby boosting the performance. Besides, since the inserted convolutions are computationally efficient, our SPViT shows higher FLOPs reduction and lower top-1 accuracy loss under large FLOPs reduction on DeiT-B.
We can also find that the SPViT-Swin variants serve as a strong baseline when pruning the hierarchical Swin~\cite{swin} models. For example, SPViT-Swin-S saves 29.9\% FLOPs with only 0.6\% Top-1 accuracy drop, outperforming STEP baseline counterparts~\cite{li2021differentiable} by 2.8\% Top-1 accuracy.

\emph{\textbf{Second, SPViT distillation variants achieve the strongest results, outperforming the SOTA methods on multiple backbones.}} We notice that the SOTA methods (UVC and MDC) achieve higher performance than the SPViT variants without knowledge distillation when pruning DeiT-Ti and DeiT-S models. The reason is that these methods enjoy a larger pruning space brought by their finer pruning granularity, as discussed in Section~\ref{subsec:related_work_pruning}. Nevertheless, our method that derives hybrid models with both MSAs and convolutions can further benefit from a CNN teacher and achieves higher performance. For example, our SPViT-DeiT-Ti$\S$, SPViT-DeiT-S$\S$, and SPViT-DeiT-B$\S$ achieve the highest top-1 accuracy among the SOTA methods of 73.2\%, 80.3\%, and 82.4\%, respectively. We also empirically find that that equipped with knowledge distillation, SPViT-Swin-Ti$\S$ and SPViT-Swin-S$\S$ achieve remarkable performance gain and barely have top-1 accuracy loss.

Furthermore, our SPViT-DeiT-B$\S$ largely outperforms token compression methods~\cite{rao2021dynamicvit,liang2022not, wei2023joint} under fewer FLOPs with DeiT-B backbone. While with DeiT-S backbone, token compression methods generally demonstrate strong performance against the module pruning methods. It's worth noting that token compression keeps the model weights intact and thus has the same or even increased model parameters. We conjecture that because DeiT-S backbone is small and more compact than DeiT-B, pruning the model weights with module pruning generally hurts the performance. Even so, our SPViT-DeiT-S$\S$ still achieves comparable FLOPs-accuracy trade-off to SOTA token compression methods dTPS~\cite{bolya2022token} and ToMe~\cite{bolya2022token}.

\subsection{Ablation Studies}
\label{subsec:ablation}
\subsubsection{Single-path vs. Multi-path Search} \label{subsec:abl_single_path}
We investigate the effectiveness of our single-path search by comparing SPViT with the multi-path counterpart in Table~\ref{tab:abl_single}. Specifically, in the multi-path implementations, we randomly initialize the weights in candidate bottleneck convolution operations before the search and keep other components the same as the single-path version. We observe that compared with multi-path counterparts, single-path search methods save more than 50\% of the search cost while obtaining better architectures with higher performance and fewer parameters under the same computational complexity. We conjecture that the superiority of our single-path formulation comes from the eased optimization difficulty and the good initialization of convolution kernels.

\subsubsection{Effectiveness of Each Module}\label{subsec:abl_each} We investigate the effectiveness of each module of our SPViT by adding them step by step when pruning DeiT-Ti and DeiT-B. The results on ImageNet-1k are reported in Table~\ref{tab:ablation}.

\subsubsection{Pruning solely MSA or FFN layers.} We investigate the effectiveness of pruning solely MSA or FFN layers under similar FLOPs savings. For both SPViT-DeiT-Ti and SPViT-DeiT-B, we empirically find that solely pruning MSA has slightly higher performance than solely pruning FFN, \ie, 0.1\% higher top-1 accuracy. However, it exhibits more parameters, However, it exhibits more parameters, as FFN pruning is more parameter-efficient, e.g., saving 0.04G FLOPs prunes 0.1M and 0.2M model parameters respectively for MSA (in both our SPViT and head pruning~\cite{michel2019sixteen}) and FFN pruning. Moreover, jointly pruning MSA and FFN layers achieves similar or better performance compared to solely pruning MSA or FFN layers while maintaining comparable or lower FLOPs. We conjecture that jointly pruning MSA and FFN layers greatly enlarges the search space, which facilitates finding powerful compact architectures with optimal MSA-FFN pruning trade-offs. We also observe that jointly pruning MSA and FFN layers has more model parameters than solely pruning FFN under the same FLOPs saving (SPViT-DeiT-Ti). This is because joint pruning needs to balance between pruning MSA and FFN layers. Nevertheless, when pruning DeiT-B with higher redundancy, we can seek more aggressive FLOPs saving in joint pruning to have a low number of parameters without sacrificing too much accuracy.

\subsubsection{Knowledge distillation.} We employ hard label distillation to compensate for the accuracy loss following the recent pruning works~\cite{yu2022unified,chen2021chasing}. We observe that it achieves even better performance than the compact models. To further investigate this phenomenon, we report the results of different teacher choices in Table~\ref{tab:abl_teacher} under the same setting of SPViT-DeiT-Ti. To better reflect the performance gain, we take UVC-DeiT-Ti$\S$~\cite{yu2022unified} as a reference, which employs the uncompressed DeiT-Ti model as the teacher and performs soft-label distillation. We find that the distillation strategy of UVC-DeiT-Ti$\S$ provides a significant 2.0\% top-1 accuracy gain. However, our SPViT-DeiT-Ti$\S$ with the same distillation strategy only has 0.4\% top-1 accuracy improvement from SPViT-DeiT-Ti. As is introduced in Section~\ref{subsec:related_work}, convolutions and MSAs exhibit different behaviors and learn different feature representations, we speculate that the knowledge from the uncompressed DeiT-Ti teacher might be noisy for the hybrid architecture searched by our SPViT. Nevertheless, when switching to a CNN teacher RegNetY-1.6GF that has similar capacity as DeiT-Ti, SPViT-DeiT-Ti$\S$ achieves a remarkable performance leap of 2.1\% top-1 accuracy, which we conjecture that the searched convolutional layers in SPViT can receive better guidance from the CNN teacher. Finally, we switch to RegNetY-16GF, a CNN teacher with higher capacity, showing that the performance can be further improved for 0.4\% top-1 accuracy, outperforming UVC-DeiT-Ti$\S$ with RegNetY-16GF by a large margin.

\begin{table}[tb!]
\caption{Effect of the teacher choices for knowledge distillation with SPViT-DeiT-Ti on ImageNet-1k. We take UVC-DeiT-Ti$\S$~\cite{yu2022unified} as a reference. $\S$ indicates fine-tuning with knowledge distillation.}\vspace{-0.8em}
\centering
\resizebox{0.4\textwidth}{!}{%
    \begin{tabular}{c|cc}
    Model & Teacher & Top-1 Acc. (\%) \\ \shline
    UVC-DeiT-Ti & - & 69.3 \\
    UVC-DeiT-Ti$\S$ & DeiT-Ti & 71.3 \\
    UVC-DeiT-Ti$\S$ & RegNetY-1.6GF~\cite{radosavovic2020designing} & 70.5 \\
    UVC-DeiT-Ti$\S$ & RegNetY-16GF~\cite{radosavovic2020designing} & 70.8 \\\hline
    SPViT-DeiT-Ti & - & 70.7 \\
    SPViT-DeiT-Ti$\S$ & DeiT-Ti & 71.1 \\
    SPViT-DeiT-Ti$\S$ & RegNetY-1.6GF~\cite{radosavovic2020designing} & 72.8 \\
    SPViT-DeiT-Ti$\S$ & RegNetY-16GF~\cite{radosavovic2020designing} & 73.2 \\
    \end{tabular}%
}
\label{tab:abl_teacher}
\end{table}

\begin{table}[t!]
\centering
\caption{Comparison with random search when only pruning MSAs and FFNs with SPViT-DeiT-Ti on ImageNet-1k.}\vspace{-0.8em}
\resizebox{0.45\textwidth}{!}{%
    \begin{tabular}{c|cc}
    Model & Top-1 Acc. (\%) & Top-5 Acc. (\%) \\ \shline
    Prune MSA only & 70.7 & 90.2 \\
    Random Search MSA & 70.2 & 90.1 \\ \hline
    Prune FFN only & 70.6 & 90.2\\
    Random Search FFN & 70.4 & 90.2
    \end{tabular}%
}
\label{tab:abl_random}
\end{table}
\subsubsection{Effect of the Search Strategy} To validate the effectiveness of our searching strategy for each module, we compare SPViT-DeiT-Ti variants with randomly searched architectures in Table~\ref{tab:abl_random}. Specifically, we explore comparisons in two settings: 1) keep the FFN MLP expansion ratios fixed, compare only pruning MSA into convolutional layers with randomly sampled MSA or convolutional layers; 2) keep the MSA layers fixed, compare only pruning FFN expansion ratios with randomly sampled FFN expansion ratios. For each setting, we randomly sample 5 architectures with computational complexity around the target FLOPs and then fine-tune these architectures and report the highest Top-1 accuracy. We observe that under the same FLOPs, our SPViT variants excel the random search counterparts by 0.5\% and 0.2\% Top-1 accuracy for the two settings, respectively. The performance gain demonstrates the effectiveness of our search strategy.

\begin{table}[tb!]
\centering
\caption{Pruning proportions for individual components. For example, MSA Saving is derived with: Pruned MSA FLOPs $/$ Total MSA FLOPs. The listed SPViT variants are the same ones as in Table~\ref{tab:swin}.}\vspace{-0.8em}
\resizebox{0.48\textwidth}{!}{%
\begin{tabular}{c|ccc}
Model & FLOPs Saving (\%) & MSA Saving (\%) & FFN Saving (\%) \\ \shline
SPViT-DeiT-Ti & 23.2 & 24.1 & 19.2 \\
SPViT-DeiT-B & 52.0 & 37.7 & 60.7 \\
SPViT-Swin-Ti & 24.4 & 33.6 & 21.3 \\
SPViT-Swin-S & 29.9 & 21.3 & 35.2 
\end{tabular}%
}
\vspace{-1.5em}
\label{tab:individual}
\end{table}

\subsubsection{Pruning Proportions for MSA and FFN Layers}\label{subsec:pruning_proportion}
As our SPViT can automatically search architectures under computational complexity constraints, we investigate the pruning proportions for MSA and FFN layers in Table~\ref{tab:individual}. We observe that the pruning proportions are different among the models, \eg, SPViT-Swin-Ti has a higher pruning ratio for MSA layers while SPViT-Swin-S has a higher pruning ratio for FFN layers. It suggests that given target efficiency constraints, SPViT can flexibly customize the suitable pruning proportions for different dense models.

\begin{figure*}[t]
  \centering
  \includegraphics[width=\linewidth]{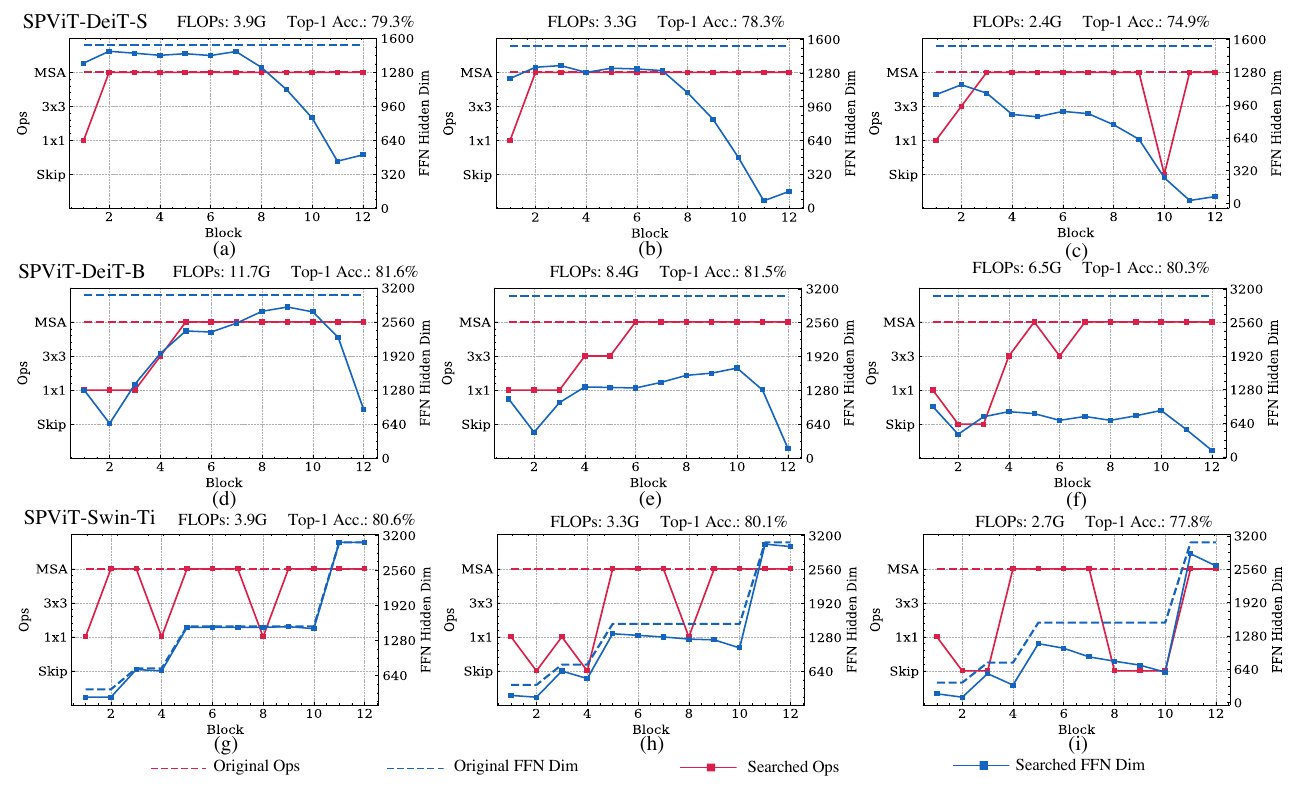}
  \vspace{-2em}
  \caption{SPViT searched architectures under different target FLOPs on ImageNet-1k. (a)-(c), (d)-(f), and (g)-(i) are architectures searched by SPViT-DeiT-S, SPViT-DeiT-B, and SPViT-Swin-Ti, respectively. Dashed lines represent the operations and the number of hidden dimensions for FFN layers before pruning. Solid lines represent the two types of architecture configurations searched by SPViT.}
  \vspace{-1em}
  \label{fig:deit_flops}
\end{figure*}

\subsubsection{Impact of Introducing Locality}\label{subsec:locality}
Our method prunes self-attentions into convolutional layers to compress the model while introducing locality. To investigate the impact of introducing locality, we keep the FFN layers intact and compare our method with two head pruning baselines under the same target FLOPs. One baseline selects the remaining heads randomly, and the other selects the remaining heads with the same NAS-based gating mechanism as SPViT. The results are reported in Table~\ref{tab:locality_vs_sparsity}. We observe that our method outperforms the baselines in Rows 2 and 3 by large margins. The results validate our motivation to introduce suitable locality in ViT pruning.

\begin{table}[tb!]
\caption{Comparisons with NAS-based and random head pruning for SPViT-DeiT-Ti on ImageNet-1k. We keep the FFN layers intact.}\vspace{-0.8em}
\resizebox{0.48\textwidth}{!}{
\centering
\begin{tabular}{c|c|cc} Row & Model & Top-1 Acc. (\%) & FLOPs (G) \\
\hline
1 & SPViT-DeiT-Ti w/o Prune FFN & 70.7 & 1.0 \\
2 & Row 1 with Random Head Pruning & 69.6 & 1.0  \\
3 & Row 1 with NAS-based Head Pruning   & 69.8 & 1.0 \\
\end{tabular}
}
\vspace{-1em}
\label{tab:locality_vs_sparsity}
\end{table}

\vspace{-1em}
\subsection{Observations on Searched Architectures}\label{subsec:observation}
Our SPViT can automatically prune pre-trained ViT models into efficient ones with proper locality inserted under desired efficiency constraints during the search. Here, we show empirical observations on the searched architectures using SPViT on standard and hierarchical ViTs when gradually decreasing the target FLOPs. In Figure~\ref{fig:deit_flops}, we visualize the searched architectures for SPViT-DeiT-S under 3.9G, 3.3G and 2.4G FLOPs, SPViT-DeiT-B under 11.7G, 8.4G, and 6.5G FLOPs, and SPViT-Swin-Ti under 3.9G, 3.4G and 2.7G FLOPs.

\begin{figure*}[t!]
  \centering
  \includegraphics[width=\linewidth]{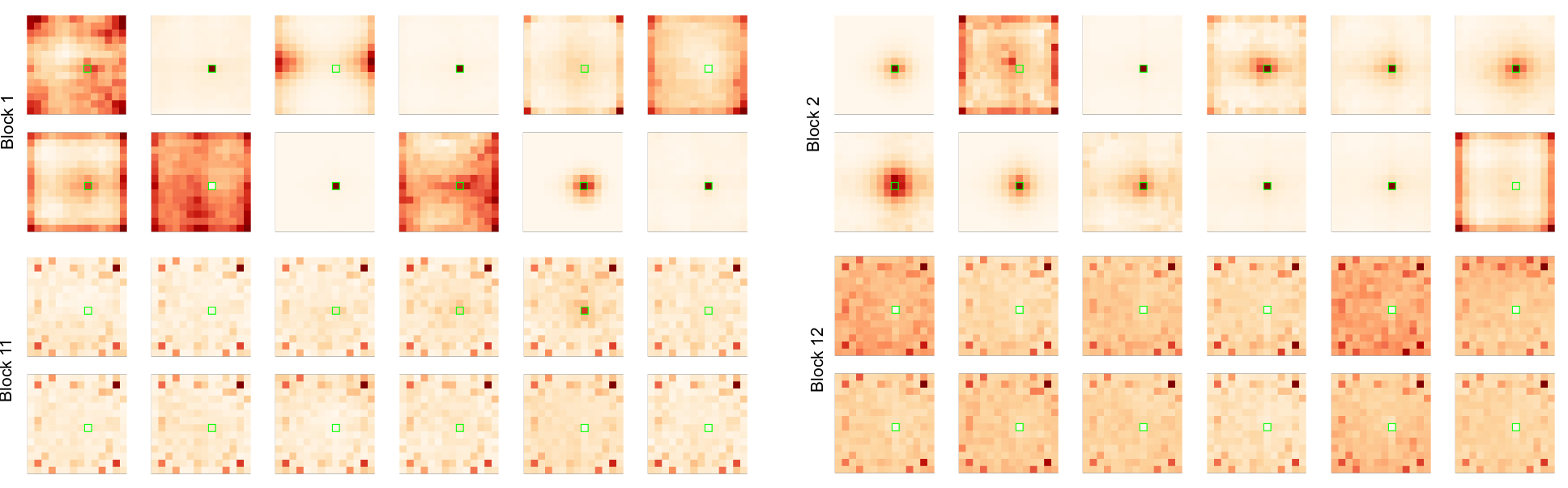}
  \vspace{-2em}
  \caption{Attention probabilities of the pre-trained DeiT-B model averaged over 100 images. We visualize the attention maps for all twelve heads in the shallow blocks (Block 1 and 2) and deep blocks (Block 11 and 12). Each map shows the attention probabilities between a query feature embedding (\emph{green square}) and the other feature embeddings. Darker color indicates higher attention probability and vice versa. Best viewed in color.}
  \vspace{-1em}
  \label{fig:heatmaps}
\end{figure*}

\subsubsection{Locality is Encouraged in Shallow Blocks}
As depicted in Figure~\ref{fig:deit_flops} (a)-(i), SPViT variants learn to prune the shallow MSA layers into bottleneck convolutional layers or skip-connections (in the first two blocks for SPViT-DeiT-S, the first six blocks for SPViT-DeiT-B, and the first three blocks for SPViT-Swin-Ti). This observation aligns with the previous work~\cite{cordonnier2019relationship,pan2021less} indicating that the shallow MSA layers contain more superfluous global correlations. The last two blocks are observed to remain as the MSA layers for all models, demonstrating the importance of the global operations in the deep layers. To further support the observation, we visualize the attention probabilities for DeiT-B pre-trained model in Figure~\ref{fig:heatmaps}. We can observe that the attention probabilities show clear different patterns between the shallow and deep blocks. In shallow blocks (Blocks 1 and 2), many MSA heads attend to the local regions around the query feature embedding, while nearly all heads in the deep blocks (Blocks 11 and 12) attend to the global regions. Therefore, the attention probability patterns suggest that our SPViT encourages pruning the shallow MSA layers into convolutional layers while keeping the deep MSA layers unchanged is reasonable.

\vspace{-0.5em}
\subsubsection{Deep FFN Layers Have More Redundancy in Standard ViTs}
For the pre-trained DeiT models, the MLP expansion ratios are set to be the same across all the blocks. As illustrated in Figure~\ref{fig:deit_flops} (a)-(f), when pruning DeiT models, the networks prioritize pruning the hidden dimensions in the last few blocks. As a higher number of hidden dimensions in FFN layers indicates a higher capacity and vice versa~\cite{lample2019large,dong2021attention}, we conjecture that the last few FFN layers need less capacity compared to the others since they handle less diverse attention maps~\cite{touvron2021going,zhou2021deepvit,goyal2020power} in standard ViTs.

\subsubsection{Shallower FFN layers Within Each Stage Require Higher Capacity in Hierarchical ViTs}
Unlike standard ViTs, the Swin models have hierarchical architectures that separate the blocks into several stages. At the beginning of each stage, the input tokens are merged into higher embedding dimensions. As shown in Figure~\ref{fig:deit_flops} (g)-(i), within the same stage, the shallower layers preserve more hidden dimensions. We conjecture that modeling the changes in the feature space caused by token merging requires a higher capacity for hierarchical ViTs.

\section{Conclusion and future work}
In this paper, we have introduced a novel weight-sharing scheme between MSA and convolutional operations, which allows expressing convolutional operations with a subset of MSA parameters and optimizing the two types of operations simultaneously. Based on the weight-sharing scheme, we have proposed SPViT to reduce the computational cost of ViTs as well as introducing proper inductive bias automatically with low search cost. Specifically, SPViT can search whether to prune the MSA into convolutional layers in ViTs and search for optimal hidden dimensions of FFN layers under desired efficiency constraints. By applying SPViT to two popular ViT variants, we have made some meaningful observations towards the importance of ViT components.

One limitation of SPViT is that the granularity for pruning MSA into convolutional layers is too coarse, where pruning the former to the latter incurs a large loss of discriminative capability due to the huge parameter gap. Hence, we will explore adding head pruning into SPViT as future work. Another future direction is to extend the weight-sharing scheme to other convolutional operation variants such as depth-wise convolution~\cite{howard2017mobilenets} and group convolution~\cite{zhang2018shufflenet}, providing broader applications.

\ifCLASSOPTIONcaptionsoff
  \newpage
\fi

\bibliographystyle{IEEEtran}
\bibliography{reference}

\begin{thebibliography}{10}
\providecommand{\url}[1]{#1}
\csname url@samestyle\endcsname
\providecommand{\newblock}{\relax}
\providecommand{\bibinfo}[2]{#2}
\providecommand{\BIBentrySTDinterwordspacing}{\spaceskip=0pt\relax}
\providecommand{\BIBentryALTinterwordstretchfactor}{4}
\providecommand{\BIBentryALTinterwordspacing}{\spaceskip=\fontdimen2\font plus
\BIBentryALTinterwordstretchfactor\fontdimen3\font minus \fontdimen4\font\relax}
\providecommand{\BIBforeignlanguage}[2]{{%
\expandafter\ifx\csname l@#1\endcsname\relax
\typeout{** WARNING: IEEEtran.bst: No hyphenation pattern has been}%
\typeout{** loaded for the language `#1'. Using the pattern for}%
\typeout{** the default language instead.}%
\else
\language=\csname l@#1\endcsname
\fi
#2}}
\providecommand{\BIBdecl}{\relax}
\BIBdecl

\bibitem{vit}
A.~Dosovitskiy, L.~Beyer, A.~Kolesnikov, D.~Weissenborn, X.~Zhai, T.~Unterthiner, M.~Dehghani, M.~Minderer, G.~Heigold, S.~Gelly, J.~Uszkoreit, and N.~Houlsby, ``An image is worth 16x16 words: Transformers for image recognition at scale,'' in \emph{ICLR}, 2021.

\bibitem{hvt}
Z.~Pan, B.~Zhuang, J.~Liu, H.~He, and J.~Cai, ``Scalable visual transformers with hierarchical pooling,'' in \emph{ICCV}, 2021.

\bibitem{swin}
Z.~Liu, Y.~Lin, Y.~Cao, H.~Hu, Y.~Wei, Z.~Zhang, S.~Lin, and B.~Guo, ``Swin transformer: Hierarchical vision transformer using shifted windows,'' in \emph{ICCV}, 2021.

\bibitem{touvron2021training}
H.~Touvron, M.~Cord, M.~Douze, F.~Massa, A.~Sablayrolles, and H.~J{\'e}gou, ``Training data-efficient image transformers \& distillation through attention,'' in \emph{ICML}, 2021, pp. 10\,347--10\,357.

\bibitem{zhang2022vsa}
Q.~Zhang, Y.~Xu, J.~Zhang, and D.~Tao, ``Vsa: Learning varied-size window attention in vision transformers,'' in \emph{ECCV}, 2022.

\bibitem{zheng2021rethinking}
S.~Zheng, J.~Lu, H.~Zhao, X.~Zhu, Z.~Luo, Y.~Wang, Y.~Fu, J.~Feng, T.~Xiang, P.~H. Torr \emph{et~al.}, ``Rethinking semantic segmentation from a sequence-to-sequence perspective with transformers,'' in \emph{CVPR}, 2021, pp. 6881--6890.

\bibitem{cheng2021per}
B.~Cheng, A.~Schwing, and A.~Kirillov, ``Per-pixel classification is not all you need for semantic segmentation,'' \emph{NeurIPS}, vol.~34, 2021.

\bibitem{wang2021max}
H.~Wang, Y.~Zhu, H.~Adam, A.~Yuille, and L.-C. Chen, ``Max-deeplab: End-to-end panoptic segmentation with mask transformers,'' in \emph{CVPR}, 2021, pp. 5463--5474.

\bibitem{carion2020end}
N.~Carion, F.~Massa, G.~Synnaeve, N.~Usunier, A.~Kirillov, and S.~Zagoruyko, ``End-to-end object detection with transformers,'' in \emph{ECCV}, 2020, pp. 213--229.

\bibitem{zhu2021deformable}
X.~Zhu, W.~Su, L.~Lu, B.~Li, X.~Wang, and J.~Dai, ``Deformable {\{}detr{\}}: Deformable transformers for end-to-end object detection,'' in \emph{ICLR}, 2021.

\bibitem{gao2021fast}
P.~Gao, M.~Zheng, X.~Wang, J.~Dai, and H.~Li, ``Fast convergence of detr with spatially modulated co-attention,'' in \emph{ICCV}, 2021, pp. 3621--3630.

\bibitem{transformer}
A.~Vaswani, N.~Shazeer, N.~Parmar, J.~Uszkoreit, L.~Jones, A.~N. Gomez, L.~Kaiser, and I.~Polosukhin, ``Attention is all you need,'' \emph{NeurIPS}, pp. 5998--6008, 2017.

\bibitem{zhu2021visual}
M.~Zhu, K.~Han, and Y.~Tang, ``Visual transformer pruning,'' in \emph{KDDW}, 2021.

\bibitem{yang2021nvit}
H.~Yang, H.~Yin, P.~Molchanov, H.~Li, and J.~Kautz, ``Nvit: Vision transformer compression and parameter redistribution,'' in \emph{CVPR}, 2023.

\bibitem{li2021localvit}
Y.~Li, K.~Zhang, J.~Cao, R.~Timofte, and L.~Van~Gool, ``Localvit: Bringing locality to vision transformers,'' \emph{arXiv preprint arXiv:2104.05707}, 2021.

\bibitem{guo2021cmt}
J.~Guo, K.~Han, H.~Wu, C.~Xu, Y.~Tang, C.~Xu, and Y.~Wang, ``Cmt: Convolutional neural networks meet vision transformers,'' in \emph{CVPR}, 2022.

\bibitem{xiao2021early}
T.~Xiao, M.~Singh, E.~Mintun, T.~Darrell, P.~Doll{\'a}r, and R.~Girshick, ``Early convolutions help transformers see better,'' in \emph{NeurIPS}, 2021.

\bibitem{xu2021vitae}
Y.~Xu, Q.~Zhang, J.~Zhang, and D.~Tao, ``Vitae: Vision transformer advanced by exploring intrinsic inductive bias,'' in \emph{NeurIPS}, 2021.

\bibitem{chen2021glit}
B.~Chen, P.~Li, C.~Li, B.~Li, L.~Bai, C.~Lin, M.~Sun, J.~Yan, and W.~Ouyang, ``Glit: Neural architecture search for global and local image transformer,'' in \emph{ICCV}, 2021, pp. 12--21.

\bibitem{xu2021bert}
J.~Xu, X.~Tan, R.~Luo, K.~Song, J.~Li, T.~Qin, and T.-Y. Liu, ``Nas-bert: Task-agnostic and adaptive-size bert compression with neural architecture search,'' in \emph{KDD}, 2021.

\bibitem{li2021bossnas}
C.~Li, T.~Tang, G.~Wang, J.~Peng, B.~Wang, X.~Liang, and X.~Chang, ``Bossnas: Exploring hybrid cnn-transformers with block-wisely self-supervised neural architecture search,'' in \emph{ICCV}, 2021.

\bibitem{cordonnier2019relationship}
J.-B. Cordonnier, A.~Loukas, and M.~Jaggi, ``On the relationship between self-attention and convolutional layers,'' in \emph{ICLR}, 2020.

\bibitem{yu2022unified}
S.~Yu, T.~Chen, J.~Shen, H.~Yuan, J.~Tan, S.~Yang, J.~Liu, and Z.~Wang, ``Unified visual transformer compression,'' in \emph{ICLR}, 2022.

\bibitem{chen2021chasing}
T.~Chen, Y.~Cheng, Z.~Gan, L.~Yuan, L.~Zhang, and Z.~Wang, ``Chasing sparsity in vision transformers: An end-to-end exploration,'' \emph{NeurIPS}, 2021.

\bibitem{molchanov2016pruning}
P.~Molchanov, S.~Tyree, T.~Karras, T.~Aila, and J.~Kautz, ``Pruning convolutional neural networks for resource efficient inference,'' in \emph{ICLR}, 2017.

\bibitem{russakovsky2015imagenet}
O.~Russakovsky, J.~Deng, H.~Su, J.~Krause, S.~Satheesh, S.~Ma, Z.~Huang, A.~Karpathy, A.~Khosla, M.~Bernstein \emph{et~al.}, ``Imagenet large scale visual recognition challenge,'' \emph{IJCV}, vol. 115, no.~3, pp. 211--252, 2015.

\bibitem{pan2021ia}
B.~Pan, Y.~Jiang, R.~Panda, Z.~Wang, R.~Feris, and A.~Oliva, ``Ia-red$^2$: Interpretability-aware redundancy reduction for vision transformers,'' \emph{NeurIPS}, 2021.

\bibitem{rao2021dynamicvit}
Y.~Rao, W.~Zhao, B.~Liu, J.~Lu, J.~Zhou, and C.-J. Hsieh, ``Dynamicvit: Efficient vision transformers with dynamic token sparsification,'' \emph{NeurIPS}, 2021.

\bibitem{bolya2022token}
D.~Bolya, C.-Y. Fu, X.~Dai, P.~Zhang, C.~Feichtenhofer, and J.~Hoffman, ``Token merging: Your vit but faster,'' in \emph{ICLR}, 2023.

\bibitem{wei2023joint}
S.~Wei, T.~Ye, S.~Zhang, Y.~Tang, and J.~Liang, ``Joint token pruning and squeezing towards more aggressive compression of vision transformers,'' in \emph{Proceedings of the IEEE/CVF Conference on Computer Vision and Pattern Recognition}, 2023, pp. 2092--2101.

\bibitem{michel2019sixteen}
P.~Michel, O.~Levy, and G.~Neubig, ``Are sixteen heads really better than one?'' \emph{NeurIPS}, vol.~32, 2019.

\bibitem{behnke2020losing}
M.~Behnke and K.~Heafield, ``Losing heads in the lottery: Pruning transformer attention in neural machine translation,'' in \emph{EMNLP}, 2020, pp. 2664--2674.

\bibitem{mao2021tprune}
J.~Mao, H.~Yang, A.~Li, H.~Li, and Y.~Chen, ``Tprune: Efficient transformer pruning for mobile devices,'' \emph{TCPS}, vol.~5, no.~3, pp. 1--22, 2021.

\bibitem{li2020efficient}
B.~Li, Z.~Kong, T.~Zhang, J.~Li, Z.~Li, H.~Liu, and C.~Ding, ``Efficient transformer-based large scale language representations using hardware-friendly block structured pruning,'' in \emph{EMNLP}, 2020.

\bibitem{bartoldson2019generalization}
B.~R. Bartoldson, A.~S. Morcos, A.~Barbu, and G.~Erlebacher, ``The generalization-stability tradeoff in neural network pruning,'' \emph{NeurIPS}, 2020.

\bibitem{chen2020lottery}
T.~Chen, J.~Frankle, S.~Chang, S.~Liu, Y.~Zhang, Z.~Wang, and M.~Carbin, ``The lottery ticket hypothesis for pre-trained bert networks,'' \emph{NeurIPS}, 2020.

\bibitem{prasanna2020bert}
S.~Prasanna, A.~Rogers, and A.~Rumshisky, ``When bert plays the lottery, all tickets are winning,'' in \emph{EMNLP}, 2020.

\bibitem{hou2021multi}
Z.~Hou and S.-Y. Kung, ``Multi-dimensional model compression of vision transformer,'' in \emph{ICME}, 2022.

\bibitem{yu2023x}
L.~Yu and W.~Xiang, ``X-pruner: explainable pruning for vision transformers,'' in \emph{CVPR}, 2023, pp. 24\,355--24\,363.

\bibitem{evci2020rigging}
U.~Evci, T.~Gale, J.~Menick, P.~S. Castro, and E.~Elsen, ``Rigging the lottery: Making all tickets winners,'' in \emph{ICML}.\hskip 1em plus 0.5em minus 0.4em\relax PMLR, 2020, pp. 2943--2952.

\bibitem{hinton2015distilling}
G.~Hinton, O.~Vinyals, and J.~Dean, ``Distilling the knowledge in a neural network,'' in \emph{NeurIPSW}, 2014.

\bibitem{geirhos2018imagenet}
R.~Geirhos, P.~Rubisch, C.~Michaelis, M.~Bethge, F.~A. Wichmann, and W.~Brendel, ``Imagenet-trained cnns are biased towards texture; increasing shape bias improves accuracy and robustness,'' in \emph{ICLR}, 2019.

\bibitem{brendel2019approximating}
W.~Brendel and M.~Bethge, ``Approximating cnns with bag-of-local-features models works surprisingly well on imagenet,'' in \emph{ICLR}, 2019.

\bibitem{naseer2021intriguing}
M.~Naseer, K.~Ranasinghe, S.~Khan, M.~Hayat, F.~S. Khan, and M.-H. Yang, ``Intriguing properties of vision transformers,'' \emph{NeurIPS}, 2021.

\bibitem{caron2021emerging}
M.~Caron, H.~Touvron, I.~Misra, H.~J\'egou, J.~Mairal, P.~Bojanowski, and A.~Joulin, ``Emerging properties in self-supervised vision transformers,'' in \emph{ICCV}, 2021.

\bibitem{mehta2021mobilevit}
S.~Mehta and M.~Rastegari, ``Mobilevit: light-weight, general-purpose, and mobile-friendly vision transformer,'' in \emph{ICLR}, 2022.

\bibitem{zhang2023lite}
N.~Zhang, F.~Nex, G.~Vosselman, and N.~Kerle, ``Lite-mono: A lightweight cnn and transformer architecture for self-supervised monocular depth estimation,'' in \emph{CVPR}, 2023, pp. 18\,537--18\,546.

\bibitem{zhang2023completionformer}
Y.~Zhang, X.~Guo, M.~Poggi, Z.~Zhu, G.~Huang, and S.~Mattoccia, ``Completionformer: Depth completion with convolutions and vision transformers,'' in \emph{CVPR}, 2023, pp. 18\,527--18\,536.

\bibitem{gao2022convmae}
P.~Gao, T.~Ma, H.~Li, Z.~Lin, J.~Dai, and Y.~Qiao, ``Convmae: Masked convolution meets masked autoencoders,'' \emph{NeurIPS}, 2022.

\bibitem{touvron2021augmenting}
H.~Touvron, M.~Cord, A.~El-Nouby, P.~Bojanowski, A.~Joulin, G.~Synnaeve, and H.~J{\'e}gou, ``Augmenting convolutional networks with attention-based aggregation,'' \emph{arXiv preprint arXiv:2112.13692}, 2021.

\bibitem{Graham_2021_ICCV}
B.~Graham, A.~El-Nouby, H.~Touvron, P.~Stock, A.~Joulin, H.~Jegou, and M.~Douze, ``Levit: A vision transformer in convnet's clothing for faster inference,'' in \emph{ICCV}, October 2021, pp. 12\,259--12\,269.

\bibitem{vasu2023fastvit}
P.~K.~A. Vasu, J.~Gabriel, J.~Zhu, O.~Tuzel, and A.~Ranjan, ``Fastvit: A fast hybrid vision transformer using structural reparameterization,'' in \emph{ICCV}, 2023.

\bibitem{lu2022bridging}
Z.~Lu, H.~Xie, C.~Liu, and Y.~Zhang, ``Bridging the gap between vision transformers and convolutional neural networks on small datasets,'' \emph{NeurIPS}, vol.~35, pp. 14\,663--14\,677, 2022.

\bibitem{chen2021x}
X.~Chen, H.~Wang, and B.~Ni, ``X-volution: On the unification of convolution and self-attention,'' \emph{arXiv preprint arXiv:2106.02253}, 2021.

\bibitem{d2021convit}
S.~d'Ascoli, H.~Touvron, M.~Leavitt, A.~Morcos, G.~Biroli, and L.~Sagun, ``Convit: Improving vision transformers with soft convolutional inductive biases,'' in \emph{ICML}, 2021.

\bibitem{d2021transformed}
S.~d'Ascoli, L.~Sagun, G.~Biroli, and A.~Morcos, ``Transformed cnns: recasting pre-trained convolutional layers with self-attention,'' \emph{arXiv preprint arXiv:2106.05795}, 2021.

\bibitem{stamoulis2019single}
D.~Stamoulis, R.~Ding, D.~Wang, D.~Lymberopoulos, B.~Priyantha, J.~Liu, and D.~Marculescu, ``Single-path nas: Designing hardware-efficient convnets in less than 4 hours,'' in \emph{ECML PKDD}, 2020.

\bibitem{guo2020single}
Z.~Guo, X.~Zhang, H.~Mu, W.~Heng, Z.~Liu, Y.~Wei, and J.~Sun, ``Single path one-shot neural architecture search with uniform sampling,'' in \emph{ECCV}, 2020, pp. 544--560.

\bibitem{cai2019once}
H.~Cai, C.~Gan, T.~Wang, Z.~Zhang, and S.~Han, ``Once-for-all: Train one network and specialize it for efficient deployment,'' in \emph{ICLR}, 2020.

\bibitem{vaswani2017attention}
A.~Vaswani, N.~Shazeer, N.~Parmar, J.~Uszkoreit, L.~Jones, A.~N. Gomez, {\L}.~Kaiser, and I.~Polosukhin, ``Attention is all you need,'' \emph{NeurIPS}, pp. 5998--6008, 2017.

\bibitem{he2016deep}
K.~He, X.~Zhang, S.~Ren, and J.~Sun, ``Deep residual learning for image recognition,'' in \emph{CVPR}, 2016, pp. 770--778.

\bibitem{pan2021less}
Z.~Pan, B.~Zhuang, H.~He, J.~Liu, and J.~Cai, ``Less is more: Pay less attention in vision transformers,'' in \emph{AAAI}, 2022.

\bibitem{nair2010rectified}
V.~Nair and G.~E. Hinton, ``Rectified linear units improve restricted boltzmann machines,'' in \emph{ICML}, 2010.

\bibitem{ioffe2015batch}
S.~Ioffe and C.~Szegedy, ``Batch normalization: Accelerating deep network training by reducing internal covariate shift,'' in \emph{ICML}, 2015, pp. 448--456.

\bibitem{raghu2021vision}
M.~Raghu, T.~Unterthiner, S.~Kornblith, C.~Zhang, and A.~Dosovitskiy, ``Do vision transformers see like convolutional neural networks?'' \emph{NeurIPS}, 2021.

\bibitem{liu2021elastic}
J.~Liu, B.~Zhuang, M.~Tan, X.~Liu, D.~Phung, Y.~Li, and J.~Cai, ``Elastic architecture search for diverse tasks with different resources,'' \emph{arXiv preprint arXiv:2108.01224}, 2021.

\bibitem{li2020differentiable}
Y.~Li, G.~Hu, Y.~Wang, T.~Hospedales, N.~M. Robertson, and Y.~Yang, ``Differentiable automatic data augmentation,'' in \emph{ECCV}, 2020, pp. 580--595.

\bibitem{ba2016layer}
J.~L. Ba, J.~R. Kiros, and G.~E. Hinton, ``Layer normalization,'' \emph{arXiv preprint arXiv:1607.06450}, 2016.

\bibitem{chen2021autoformer}
M.~Chen, H.~Peng, J.~Fu, and H.~Ling, ``Autoformer: Searching transformers for visual recognition,'' in \emph{ICCV}, 2021, pp. 12\,270--12\,280.

\bibitem{liao2021searching}
Y.-L. Liao, S.~Karaman, and V.~Sze, ``Searching for efficient multi-stage vision transformers,'' \emph{arXiv preprint arXiv:2109.00642}, 2021.

\bibitem{hendrycks2016gaussian}
D.~Hendrycks and K.~Gimpel, ``Gaussian error linear units (gelus),'' \emph{arXiv preprint arXiv:1606.08415}, 2016.

\bibitem{liang2022not}
Y.~Liang, C.~Ge, Z.~Tong, Y.~Song, J.~Wang, and P.~Xie, ``Not all patches are what you need: Expediting vision transformers via token reorganizations,'' in \emph{ICLR}, 2022.

\bibitem{li2021differentiable}
J.~Li, R.~Cotterell, and M.~Sachan, ``Differentiable subset pruning of transformer heads,'' \emph{TACL}, 2021.

\bibitem{wang2021pyramid}
W.~Wang, E.~Xie, X.~Li, D.-P. Fan, K.~Song, D.~Liang, T.~Lu, P.~Luo, and L.~Shao, ``Pyramid vision transformer: A versatile backbone for dense prediction without convolutions,'' in \emph{ICCV}, 2021.

\bibitem{liu2021lbs}
J.~Liu, B.~Zhuang, P.~Chen, Y.~Guo, C.~Shen, J.~Cai, and M.~Tan, ``Single-path bit sharing for automatic loss-aware model compression,'' \emph{TPAMI}, 2023.

\bibitem{li2021dynamic}
C.~Li, G.~Wang, B.~Wang, X.~Liang, Z.~Li, and X.~Chang, ``Dynamic slimmable network,'' in \emph{CVPR}, 2021, pp. 8607--8617.

\bibitem{bengio2013estimating}
Y.~Bengio, N.~L{\'e}onard, and A.~Courville, ``Estimating or propagating gradients through stochastic neurons for conditional computation,'' \emph{arXiv preprint arXiv:1308.3432}, 2013.

\bibitem{radosavovic2020designing}
I.~Radosavovic, R.~P. Kosaraju, R.~Girshick, K.~He, and P.~Doll{\'a}r, ``Designing network design spaces,'' in \emph{CVPR}, 2020, pp. 10\,428--10\,436.

\bibitem{lample2019large}
G.~Lample, A.~Sablayrolles, M.~Ranzato, L.~Denoyer, and H.~J{\'e}gou, ``Large memory layers with product keys,'' \emph{NeurIPS}, 2019.

\bibitem{dong2021attention}
Y.~Dong, J.-B. Cordonnier, and A.~Loukas, ``Attention is not all you need: Pure attention loses rank doubly exponentially with depth,'' in \emph{ICML}, 2021.

\bibitem{touvron2021going}
H.~Touvron, M.~Cord, A.~Sablayrolles, G.~Synnaeve, and H.~J{\'e}gou, ``Going deeper with image transformers,'' in \emph{ICCV}, 2021.

\bibitem{zhou2021deepvit}
D.~Zhou, B.~Kang, X.~Jin, L.~Yang, X.~Lian, Z.~Jiang, Q.~Hou, and J.~Feng, ``Deepvit: Towards deeper vision transformer,'' \emph{arXiv preprint arXiv:2103.11886}, 2021.

\bibitem{goyal2020power}
S.~Goyal, A.~R. Choudhury, S.~Raje, V.~Chakaravarthy, Y.~Sabharwal, and A.~Verma, ``Power-bert: Accelerating bert inference via progressive word-vector elimination,'' in \emph{ICML}, 2020, pp. 3690--3699.

\bibitem{howard2017mobilenets}
A.~G. Howard, M.~Zhu, B.~Chen, D.~Kalenichenko, W.~Wang, T.~Weyand, M.~Andreetto, and H.~Adam, ``Mobilenets: Efficient convolutional neural networks for mobile vision applications,'' \emph{arXiv preprint arXiv:1704.04861}, 2017.

\bibitem{zhang2018shufflenet}
X.~Zhang, X.~Zhou, M.~Lin, and J.~Sun, ``Shufflenet: An extremely efficient convolutional neural network for mobile devices,'' in \emph{CVPR}, 2018, pp. 6848--6856.

\end{thebibliography}

\vspace{-3em}
\begin{IEEEbiography}
[{\includegraphics[width=1in,height=1.25in,clip,keepaspectratio]{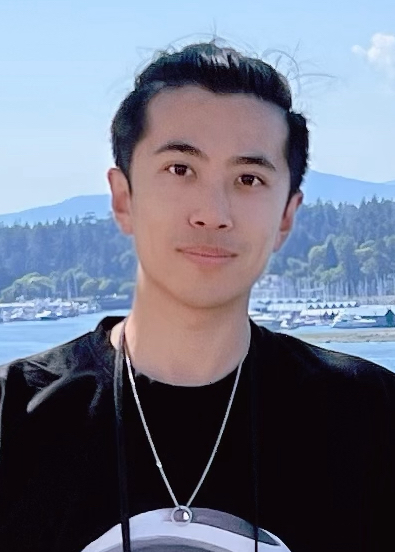}}]{Haoyu He}
is a Ph.D. student in the Faculty of Information Technology, Monash University Clayton Campus, Australia. He received his BCS and Mphil Degrees in 2019 and 2021, both from the University of Sydney, Australia.
His research interests include computer vision, efficient deployment of large models, model compression, and segmentation tasks.
\end{IEEEbiography}

\vspace{-1.5em}
\begin{IEEEbiography}[{\includegraphics[width=1in,height=1.25in,clip,keepaspectratio]{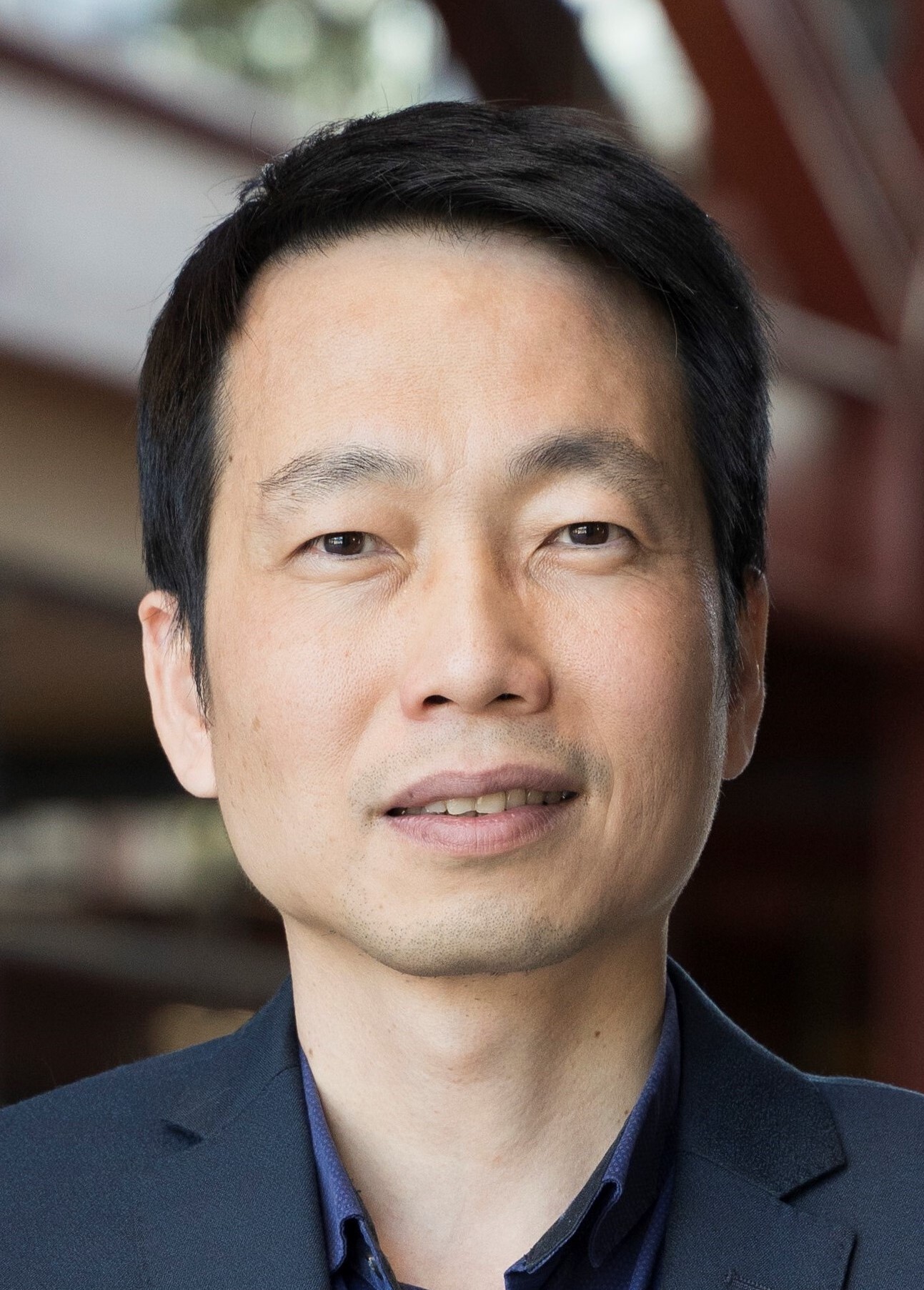}}]{Jianfei Cai}(S'98-M'02-SM'07-F’21) received his PhD degree from the University of Missouri-Columbia. He is currently a Professor and serves as the Head of the Data Science \& AI Department at Faculty of IT, Monash University, Australia. Before that, he had served as Head of Visual and Interactive Computing Division and Head of Computer Communications Division in Nanyang Technological University (NTU). His major research interests include computer vision, multimedia and visual computing. He is a co-recipient of paper awards in ACCV, ICCM, IEEE ICIP and MMSP. He serves or has served as an Associate Editor for TPAMI, IJCV, IEEE T-IP, T-MM, and T-CSVT as well as serving as Area Chair for CVPR, ICCV, ECCV, IJCAI, ACM Multimedia, ICME and ICIP. He was the Chair of IEEE CAS VSPC-TC during 2016-2018. He is the leading TPC Chair for IEEE ICME 2012 and the leading general chair for ACM Multimedia 2024. He is a Fellow of IEEE.
\end{IEEEbiography}

\vspace{-1.5em}
\begin{IEEEbiography}[{\includegraphics[width=1in,height=1.25in,clip,keepaspectratio]{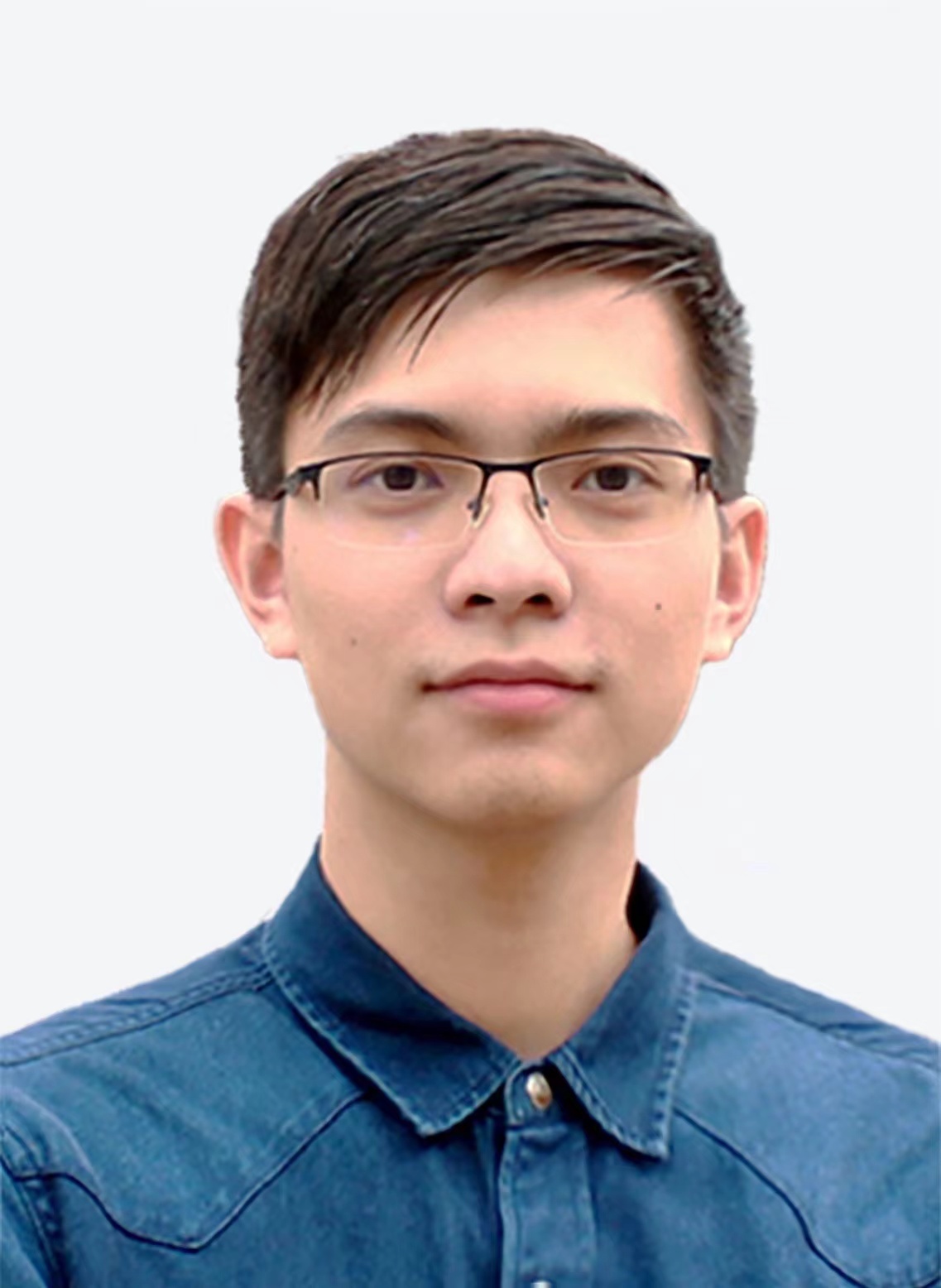}}]{Jing Liu}
is a Ph.D. student in the Faculty of Information Technology, Monash University Clayton Campus, Australia. He received his Bachelor Degree in 2017 and Master Degree in 2020, both from the School of Software Engineering at South China University of Technology, China.
His research interests include computer vision, model compression and acceleration.
\end{IEEEbiography}

\vspace{-1.5em}
\begin{IEEEbiography}[{\includegraphics[width=1in,height=1.25in,clip,keepaspectratio]{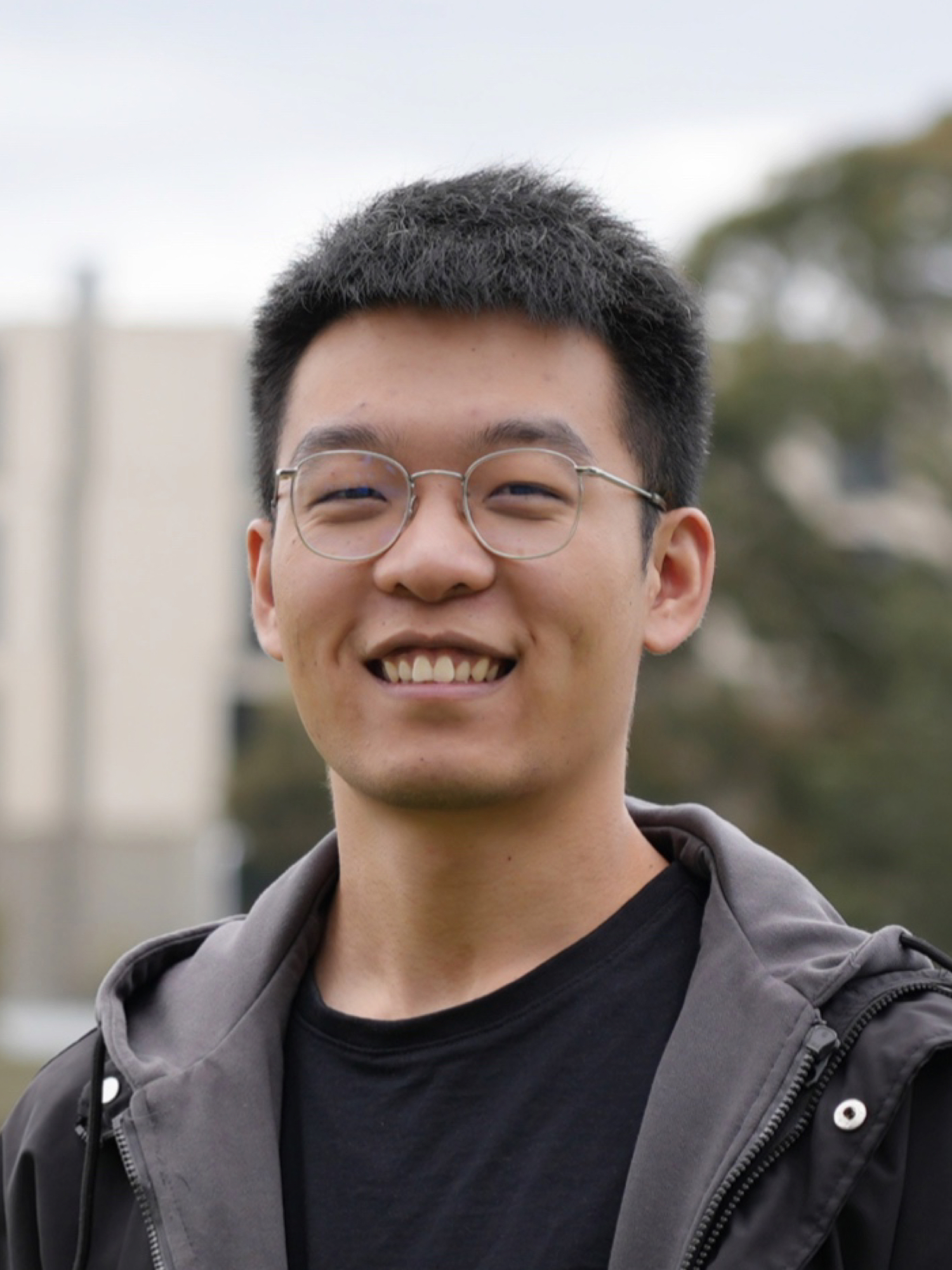}}]{Zizheng Pan} is a Ph.D. student at Department of Data Science and AI, Monash University Clayton Campus, Australia. He received his bachelor degree from Harbin Institute of Technology, Weihai Campus in 2019. Besides, he obtained his master degree from The University of Adelaide in 2020. His research topic ranges from model efficiency and scalable vision problems.
\end{IEEEbiography}

\vspace{-1.5em}
\begin{IEEEbiography}[{\includegraphics[width=1in,height=1.25in,clip,keepaspectratio]{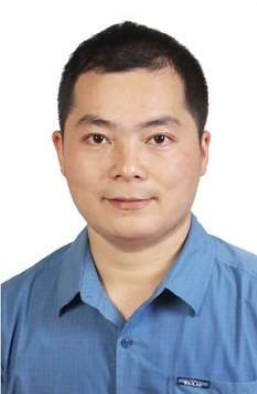}}]{Jing Zhang} (Senior Member, IEEE) is currently a Research Fellow at the School of Computer Science, The University of Sydney. He has published more than 60 papers in prestigious conferences and journals, such as CVPR, ICCV, ECCV, NeurlPS, ICLR, IEEE TPAMI, and IJCV. His research interests include computer vision and deep learning. He is also a Senior Program Committee Member of the AAAI Conference on Artificial Intelligence and the International Joint Conference on Artificial Intelligence. He serves as a regular reviewer for many prestigious journals and conferences.
\end{IEEEbiography}

\vspace{-1.5em}
\begin{IEEEbiography}
[{\includegraphics[width=1in,height=1.25in,clip,keepaspectratio]{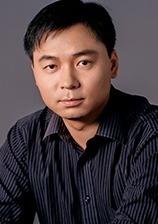}}]{Dacheng Tao} is currently a Distinguished University Professor in the School of Computer Science and Engineering at Nanyang Technological University. He mainly applies statistics and mathematics to artificial intelligence and data science, and his research is detailed in one monograph and over 200 publications in prestigious journals and proceedings at leading conferences, with best paper awards, best student paper awards, and test-of-time awards. His publications have been cited over 112K times and he has an h-index 160+ in Google Scholar. He received the 2015 and 2020 Australian Eureka Prize, the 2018 IEEE ICDM Research Contributions Award, and the 2021 IEEE Computer Society McCluskey Technical Achievement Award. He is a Fellow of the Australian Academy of Science, AAAS, ACM, and IEEE. \end{IEEEbiography}

\vspace{-1.5em}
\begin{IEEEbiography}[{\includegraphics[width=1in,height=1.25in,clip,keepaspectratio]{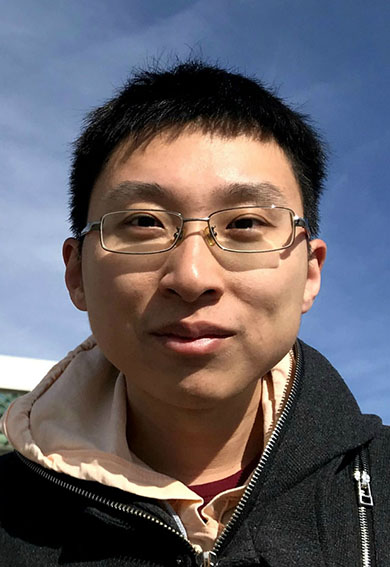}}]{Bohan Zhuang} is an assistant professor with the Faculty of Information Technology, Monash University, Australia.
He primarily focuses on
efficient machine learning research, with a particular emphasis on model quantization and
pruning, as well as designing lightweight neural architectures. He has published over 40
papers in top-tier international conferences (e.g., CVPR, NeurIPS) and journals (e.g.,
TPAMI) in the computer vision and machine learning venues.
\end{IEEEbiography}

\end{document}